\pdfoutput=1

\documentclass[11pt]{article}
\usepackage[final]{acl}

\usepackage[T1]{fontenc}

\usepackage[utf8]{inputenc}
\DeclareUnicodeCharacter{0301}{\'{}} 

\usepackage{microtype}
\usepackage{inconsolata}

\usepackage{colortbl, hhline}
\usepackage{multirow}
\usepackage[font=small,labelfont=bf]{caption}
\usepackage{amsmath,amsfonts}
\usepackage{array}
\usepackage{algorithm}
\usepackage{algpseudocode}

\usepackage{textcomp}
\usepackage{stfloats}
\usepackage{url}
\usepackage{verbatim}
\usepackage{graphicx}
\usepackage{rotating}
\usepackage{subfigure}
\usepackage{svg}
\usepackage{balance}
\usepackage{float}
\usepackage[section]{placeins}
\usepackage{booktabs,makecell,tabularx}
\usepackage{lipsum} 
\usepackage[utf8]{inputenc}
\usepackage{siunitx}
\usepackage{diagbox}
\usepackage{slashbox}
\usepackage{arydshln} 
\usepackage{afterpage}
\usepackage{pifont} 
\newcommand{\checkboxcmark}{\fbox{\ding{51}}} 
\newcommand{\checkboxempty}{\fbox{\phantom{\ding{51}}}} 

\usepackage{pgfplots, xcolor}
\usepackage{pgfplotstable} 
\pgfplotsset{compat=1.17}
\usepackage{pgfplotstable}
\usepackage{tikz}
\usetikzlibrary{pgfplots.colormaps}

\newlength\myheight
\newlength\mydepth
\settototalheight\myheight{Xygp}
\title{Revealing the impact of synthetic native samples and multi-tasking strategies in Hindi-English code-mixed humour and sarcasm detection}

\author{
Debajyoti Mazumder,  Aakash Kumar\thanks{\text{ }This work was done during his stay at Indian Institute of Science Education and Research, Bhopal.},  Jasabanta Patro \\
  Department of Data Science and Engineering \\
  Indian Institute of Science Education and Research, Bhopal, India \\
  \texttt{\{debajyoti22, aakashk19, jpatro\}@iiserb.ac.in}\\
}

\begin{document}
\maketitle

\begin{abstract}

In this paper, we reported our experiments with various strategies to improve code-mixed humour and sarcasm detection. Particularly, we tried three approaches: (i) native sample mixing, (ii) multi-task learning (MTL), and (iii) prompting and instruction finetuning very large multilingual language models (VMLMs). In native sample mixing, we added monolingual task samples to code-mixed training sets. In MTL learning, we relied on native and code-mixed samples of a semantically related task (hate detection in our case). Finally, in our third approach, we evaluated the efficacy of VMLMs via few-shot context prompting and instruction finetuning. Some interesting findings we got are (i) adding native samples improved humor (raising the F1-score up to \textbf{6.76\%}) and sarcasm (raising the F1-score up to \textbf{8.64\%}) detection, (ii) training MLMs in an MTL framework boosted performance for both humour (raising the F1-score up to \textbf{10.67\%}) and sarcasm (increment up to \textbf{12.35\%} in F1-score) detection, and (iii) prompting and instruction finetuning VMLMs couldn't outperform the other approaches. Finally, our ablation studies and error analysis discovered the cases where our model is yet to improve. We provided our code\footnote{\url{https://github.com/islnlp/code-mix-humor-sarcasm-detection-EMNLP-2025}} for reproducibility.

\end{abstract}

\section{Introduction:}

Humour and sarcasm are complex and subjective emotions that impact the nature of human communication. They can appear in different forms such as exaggeration, dark humour, gross humour, adult or slang expression, insult, offence, etc. \cite{frenda2018role, ahuja2019computational}. Past study \cite{bleakley2023politics} highlighted how they can affect politics amid tragedy. Detecting Humour and Sarcasm becomes more challenging in a code-mixed setting. This is because models now need to understand humour and sarcasm in an utterance expressed through altering multiple languages. More details on the phenomenon of code-mixing are presented in the Appendix. An example of humorous and sarcastic expression in Hindi-English code-mixed language is given in the following. More examples are presented in Figure \ref{fig:dataset_samples_table} of Appendix \ref{dataset_details}. In the following example, the English parts are marked in red, and the Hindi parts are marked in blue. We have provided their translations for readability. 


\begin{itemize}
    \item \textbf{Humor:} \textcolor{red}{Never take a moral high ground}. \textcolor{blue}{Wahan} \textcolor{red}{railing} \textcolor{blue}{nahi hai aur kabhi bhi gir sakte hain}. \\
    (\textbf{Translation:} Never take a moral high ground. There are no railings and one can fall at any time.)
    \item \textbf{Sarcasm:} \textcolor{blue}{Kuch logo ka} \textcolor{red}{number} \textcolor{blue}{iss liye} \textcolor{red}{save} \textcolor{blue}{krte hain ki galti se uth naa jaye}.... \#\textcolor{red}{sarcasm} \\
    (\textbf{Translation:} Some people save their numbers so that they don't get called by mistake.... \#sarcasm)
\end{itemize}

The NLP community has shown significant interest in monolingual humour and sarcasm detection \cite{abulaish2020survey, joshi2017automatic, joshi-etal-2020-state}. Unfortunately, there is relatively less focus on the code-mixed settings \cite{singh2023survey, elayan2022you, chen2024survey}. Therefore, we have few publicly annotated code-mixed corpora available, further acting as a bottleneck in developing new models \cite{sitaram2019survey, dogruoz-etal-2021-survey, winata-etal-2023-decades}. The evolution of multilingual large language models (MLM hereafter) has shown a new path to address this issue. They can learn task-specific knowledge from samples in one language and make predictions for samples in different languages. This phenomenon is known as cross-lingual learning. It is very effective if training and testing samples share similar linguistic and cultural contexts \cite{bigoulaeva2021cross, gupta2022multilingual}. MLMs learn their embeddings from a corpora spanning multiple languages, thus they are aware of vocabulary of multiple languages. In the context of code-mixed languages, it means we can fine-tune the MLMs using native monolingual task samples (e.g. for Hindi and English task samples for a Hindi-English code-mixed task) and do prediction for code-mixed samples. The hypothesis is that since code-mixed corpora (in Hindi-English code-mixed language) and the native monolingual corpora (in Hindi and English languages) are likely to share similar linguistic and cultural contexts, the training with native task samples or adding them in code-mixed training sets can improve code-mixed task performance. In fact, \citet{mazumder2024improving}, in a set of empirical experiments, have shown that adding Hindi and English hate samples in code-mixed hate training corpora improves code-mixed hate detection \cite{mazumder2024improving}. However, nobody has tested it for code-mixed Humour and Sarcasm detection. A detailed discussion of prior works in this direction is presented in the Appendix \ref{section:related_works}. Further, we observed that Hindi humour and sarcasm datasets are not publicly available. We experimented with synthetic Hindi samples and multi-tasking strategies to fill this gap. Overall, we asked for three research questions,

\noindent \textbf{R1:} \textit{Does  mixing native samples (English and synthetic Hindi samples in our case) in code-mixed training sets improve code-mixed humour and sarcasm detection?} \\
\textbf{R2:} \textit{Do jointly training with a semantically related third task (hate detection) along with native sample mixing improve code-mixed humour and sarcasm detection?} \\
\textbf{R3:} \textit{Do adding native samples in the prompting context or in instruction finetuning of Very large MLMs (VMLMs hereafter) improve the performance?}


\noindent In summary, our contributions are the following,
\begin{itemize}
    \item We analyzed the effect of adding native samples ( both English and synthetic Hindi samples) from existing humour and sarcasm datasets to code-mixed training data. For this, we experimented with two types of models: (i) statistical classifiers on top of word n-gram features, (ii) MLMs such as mBERT \cite{devlin-etal-2019-bert}, XLM-R \cite{xlm-r}, MuRIL \cite{khanuja2021muril} and IndicBERT \cite{doddapaneni-etal-2023-towards} (refer \textbf{Exp. 1}: Section \ref{subsection:NSM}). Combining native samples with code-mixed data led to improvements in MLMs, achieving increment up to \textbf{6.76\%} and \textbf{8.64\%} for humor and sarcasm detection, respectively ($p<0.05$). In contrast, statistical models performed worse.
    \item We integrated a related task which is hate detection with native samples in a multitask learning framework (refer \textbf{Exp. 2}: Section \ref{subsection:MTL}). Instead of sequential training, the model processed batches containing mixed-task samples. We conducted an ablation study to understand the role of the gating mechanism in the MTL framework (in Appendix \ref{section:Ablation Study}). This approach gave significant performance improvements for MLMs, with F1-score up to \textbf{10.67\%} increment in humour and \textbf{12.35\%} increment in sarcasm ($p<0.05$). With a gating mechanism, they better handled \textit{shorter contexts} and \textit{misspelt samples}. 
    \item We compared the performance of VMLMs using native and code-mixed examples as few-shots (refer \textbf{Exp. 3}: Section \ref{subsection:VMLM}). In the second set-up, we compared the VMLMs' performance when native samples are addded to the code-mixed training set. However, neither type of set-ups could improve the VMLM predictions.
\end{itemize}

\section{Datasets:}

In this section, we reported the details of code-mixed and native datasets considered in our study. Please note that we ignored the datasets containing dialogues and multi-modal samples to simplify our task formulation. The list of the datasets and their basic statistical details were reported in Table \ref{tab:dataset_stat_orig}. The dataset examples were illustrated in Figure \ref{fig:dataset_samples_table} in the appendix. A brief description of the individual datasets were reported in Appendix \ref{dataset_details}.  Apart from the class distribution, we also provided the Kullback-Leibler (KL) divergence \cite{Kullback1951} values for the individual datasets in Table \ref{tab:dataset_stat_orig}. The KL divergence measures the difference between two probability distributions. Here, it quantifies lexical variation, i.e., \textit{how the word distribution in class differs from the other}. A higher value of KL divergence suggests that the classes are well-separated in terms of word distributions. We also reported the fraction of total samples containing English hurtful (offensive, aggressive and hateful) keywords in each class. We used the lexicon given by \citet{bassignana2018hurtlex} to calculate it. Finally, we also reported the inter-annotator agreement (IAA) scores published with individual datasets. 


\begingroup
\renewcommand{\arraystretch}{1.2} 
\begin{table*}
    \centering
    \setkeys{Gin}{keepaspectratio}
    \resizebox*{\textwidth}{1.0\textheight} {
    \begin{tabular}{|l|l|l|c|c|c|c|c|c|c|}
    \hline
        ~ & Language & Dataset & \# $+ve$ & \# $-ve$ & H $+ve$ & H $-ve$ & KL & IAA \\ \hline \hline
        \multirow{5}{4em}{Humor} & Code-mixed & \citet{khandelwal-etal-2018-humor} & 1759 & 1192 & 31.32\% & 22.98\% & 1.603 & H: 0.821 \& NH: 0.794 (Fleiss’ Kappa) \\ 
        ~ & English & Col\cite{annamoradnejad2024colbert} & 100000 & 100000 & 64.92\% & 39.85\% & 2.386 & N/A \\ 
        ~ & English & POTD\cite{yang-etal-2015-humor} & 2423 & 2403 & 47.13\% & 53.05\% & 1.318 & N/A \\ 
        ~ & English & HaHa\cite{meaney-etal-2021-semeval} & 6179 & 3821 & 73.54\% & 60.64\% & 1.489 & 0.736 (Krippendorff’s) \\ 
        ~ & English & 16000\cite{mihalcea-strapparava-2005-making} & 16000 & 16000 & 49.94\% & 52.55\% & 1.209 & N/A \\  \hline

        \multirow{4}{4em}{Sarcasm} & Code-mixed & \citet{swami2018corpus} & 504 & 4746 & 31.15\% & 38.33\% & 3.021 &  0.79 (Cohen’s Kappa) \\ 
        ~ & English & NHD\cite{misra2019sarcasm} & 11724 & 14985 & 44.20\% & 39.26\% & 1.742 & N/A \\
        ~ & English & iSarc\cite{oprea-magdy-2020-isarcasm} & 1067 & 4668 & 60.44\% & 51.82\% & 1.184 & N/A \\ 
        ~ & English & SC-V2\cite{oraby-etal-2016-creating} & 4693 & 4693 & 77.69\% & 79.90\% & 0.645 & 0.80 \\  \hline

        \multirow{3}{4em}{Hate} & Code-mixed & \citet{bohra-etal-2018-dataset} & 1661 & 2914 & 71.82\% & 77.62\% &  1.218 & 0.982 (Cohen's Kappa) \\ 
        ~ & Hindi & HCHIn\cite{das2022hatecheckhin} & 3338 & 1416 & - & - & 0.740 & 0.95 (Fleiss' Kappa) \\ 
        ~ & English & HASOC\footref{refnote} & 2261 & 3591 & 76.47\% & 60.68\% & 1.182 & 72\% (overlap) \\ \hline
    \end{tabular}}

    \caption{Dataset statistics. Notation: \# for number of samples, H denotes proportion of samples containing hurtful (offensive, aggressive and hateful) keywords from positive or negative class, IAA for Inter-Annotator Agreement and KL for symmetrized smoothed Kullback-Leibler divergence between word distributions of positive ($+ve$) and negative ($-ve$) samples.}
\label{tab:dataset_stat_orig}
\end{table*}
\endgroup

\section{Experiments:}

In this section, we reported the details of our experiments conducted as a part of this study. Apart from reproducing the baselines (section \ref{exp:baselines}), we designed three experiments each based on a unique research philosophy. The three philosophies are i) native sample mixing (section \ref{subsection:NSM}), ii) multi-task learning (section \ref{subsection:MTL}) and iii) prompting and instruction finetuning very large multilingual language models (VMLMs) (section \ref{subsection:VMLM}). While the native sample mixing strategy intends to improve the code-mixed tasks by adding monolingual native samples to the code-mixed training sets, the multi-task learning strategy tries to do the same by learning linguistic knowledge from the samples of a third task (here, it is hate detection). Our last strategy, i.e. prompting and instruction finetuning VMLMs, evaluates the performance of very large multilingual language models for the considered code-mixed tasks in a few-shot context prompting and instruction finetuning scenarios. The detailed experimental set up is given in Appendix \ref{experimental_config}.


\subsection{Baselines:}
\label{exp:baselines}
In this section, we reported the previously proposed best performing methods as baselines. They were proposed for code-mixed humour and sarcasm detection in Hindi-English code-mixed scenario. Please note that some baseline papers did not share their code; thus, we reimplemented them to the best of our knowledge. Further, we made our codebase public for reproducibility. In the results section, we reported both the reproduced and original results as per the baseline paper. However, we considered the reproduced results for comparative analysis. \\
\textbf{Humor:}
We considered two previous works published by \citet{agarwal-narula-2021-humor-generation} and \citet{muttaraju-etal-2022-semi-supervised} as our baselines. Further details of individual approaches are provided in Appendix - \ref{addn_baseline_details_humour}. \\
\textbf{Sarcasm:}
We considered two previous works published by \citet{pandey2023bert} and \citet{aloria2023hilarious} as our baselines. Further details of individual approaches are provided in Appendix - \ref{addn_baseline_details_sarcasm}.

\subsection{Exp. 1- Impact of mixing native language samples:}
\label{subsection:NSM}
Our first experiment explored the impact of native language samples after adding them to the code-mixed training sets. Past study \cite{mazumder2024improving} reported that this strategy works for code-mixed hate detection. However, no one tested it for code-mixed humour and sarcasm detection. Further, in our case, even though there are publicly available English humour and sarcasm datasets (Table \ref{tab:dataset_stat_orig}), we couldn't find any Hindi datasets for the same. Thus, we choose to create silver annotated datasets by translating some portions of English datasets into Hindi using Google Translator API\footnote{\url{https://cloud.google.com/translate?hl=en}}. Note that accurately translating humour and sarcasm samples is still an open research topic. Thus, we used the most popular publicly available translation tool of current time. From the methodological point of view, we considered two types of models. 

\begin{itemize}
    \item \textbf{Statistical classifiers:} We considered three statistical classifiers, i.e. Naive Bayes (NB), Random Forest (RF), and Support Vector Machine (SVM) in our study. NB is known to perform better when there is high KL divergence between classes. We utilized word-level unigrams, bigrams, and trigrams as features. Past works showed these features to perform best \cite{khandelwal-etal-2018-humor, swami2018corpus}.
    
    \item \textbf{Multilingual Language Models (MLMs):} We also considered four widely used MLMs for our study. They are, i) mBERT \cite{devlin-etal-2019-bert}, ii) XLM-R \cite{xlm-r}, iii) MuRIL \cite{khanuja2021muril} and iv) IndicBERT \cite{doddapaneni-etal-2023-towards}. Out of them, mBERT and XLM-R are general-purpose MLMs trained on 100+ languages, while MuRIL and IndicBERT are specialized models specifically trained for Indic languages. We froze all but the last four layers of MLMs during fine-tuning.

\end{itemize}

\subsection{Exp. 2- Multi-task learning:}
\label{subsection:MTL}
In our second approach, we explored the efficiency of multi-task framework to detect code-mixed humour and sarcasm by learning linguistic knowledge from the samples of a third task, i.e., here it is, hate detection. We chose hate detection as a third task because (i) it is semantically related to humour and sarcasm, (ii) samples of humour and sarcasm datasets contained hateful keywords (refer Table \ref{tab:dataset_stat_orig}), and (iii) we could found Hindi, English and code-mixed samples available for this task. Our multi-task framework is inspired by the framework proposed by \citet{rotman-reichart-2022-multi}. 
They utilized a BERT-like architecture divided into two parts. The bottom module consists of eight lower layers of a transformer-based language model like BERT were common to the participating tasks. The top module, containing top four layers, were separately present for the individual tasks. Apart from that there is an additional top module present for parameter sharing. A gating mechanism connects all of the task-specific top modules with the additional top module. Authors experimented this framework on tasks like dependency parsing and named entity recognition and found that it performs better. In our case (i) we added a regularization term for soft parameter sharing of the final layers from the top module, and (ii) we froze the parameters of the bottom module during fine-tuning. The architecture of our framework is shown in Figure \ref{fig:MTL_arch}. We used embeddings from three widely used MLMs, i.e., (i) mBERT, (ii) XLM-R and (iii) MuRIL to initialize the layers of our MTL framework. We conducted an ablation study (refer Appendix \ref{section:Ablation Study}) to examine the role of the gating mechanism within the MTL models by removing the gate and comparing the performance with models that included the gating mechanism.

\subsection{Exp. 3- Impact of in-context learning on very large multilingual language models: }
\label{subsection:VMLM}

In our third experiment, we evaluated the performance of very large multilingual language models in detecting humour and sarcasm in code-mixed texts. As their name suggests, they have a lot of parameters, thus, they are designed to work well with prompting approaches rather than fine-tuning them fully. We utilized four VMLMs: (i) Gemma \cite{team2024gemma}, (ii) Aya Expanse \cite{ustun-etal-2024-aya}, (iii) Llama 3.1 \cite{dubey2024llama} and (iv) GPT-4\footnote{\url{https://chatgpt.com/}} \cite{achiam2023gpt}. These VMLMs were chosen for their strong performance in both Indic languages \cite{watts2024pariksha} and English. We used two scenarios: (i) few-shot prompting and (ii) instruction finetuning using LoRA \cite{hu2021lora} adapter. Here in the few shot set-up, we show a few training examples in the prompt while asking the VMLMs to classify for a test sample. This strategy has proven its superiority in sentiment-related code-mixed tasks\cite{yadav2024leveraging}. The prompt template and some of the examples are reported in Appendix \ref{section:prompts}. We conducted many variants of this experiment by considering samples from native and code-mixed training set as few shots examples in the prompt. In the second scenario, we finetuned the first three open source VMLMs on the code-mixed training set and then after combining the native language samples with the code-mixed training data.

\section{Results and discussion:}

As the datasets are not balanced, we evaluated our models and the baselines using the F1-score(\textit{F1}). The reported values are the average of three random seeds over separate runs. In the following subsections, we reported the results of our baseline and three experiments. Statistically significant results ($p<0.05$) are identified with an `*' mark.


\subsection{Baseline results:}

Most of the baseline papers reported accuracy values as their performance measures. However, since the datasets are class-imbalanced, we believe F1-scores best measures their performance. We implemented the baselines and reported the F1-scores in Table \ref{tab:results_baselines}. Out of all methods, we found that IndicBERT performed best for humour detection. Similarly, LSTM-BERT gave the best result for sarcasm detection.  


\begingroup
\renewcommand{\arraystretch}{1.08} 
\begin{table}
    \centering
    \setkeys{Gin}{keepaspectratio}
    \resizebox{\columnwidth}{\textheight} {
    \begin{tabular}{l|l|c}
    \hline
        ~ & Baselines & F1 \\ \hline \hline  
        \multirow{2}{4em}{Humor} & \citet{agarwal-narula-2021-humor-generation}  & $\textbf{0.78}^\dag$ \\
        & \citet{muttaraju-etal-2022-semi-supervised} & $0.74^\dag$ \\ \hline
        \multirow{2}{4em}{Sarcasm} & \citet{pandey2023bert} & $\textbf{0.85}^\dag$ ($0.92^{\#}$) \\ 
        & \citet{aloria2023hilarious} & $0.84^\dag$ ($0.84^{\#}$) \\ \hline
    \end{tabular}}
    \caption{Baselines results. Notation: originally reported scores are mentioned with `$\#$' mark and our reproduced results are marked as `\dag'.}
    \label{tab:results_baselines}
\end{table}
\endgroup



\subsection{Observations from Exp. 1:}
\label{Exp. 1 Results}

\begingroup
\renewcommand{\arraystretch}{1.15} 
\begin{table*}
    \centering
    \setkeys{Gin}{keepaspectratio}
    \resizebox*{\textwidth}{0.99\textheight} {
    \begin{tabular}{>{\cellcolor{white}}l|>{\cellcolor{white}}cccc>{\cellcolor{white}}c>{\cellcolor{white}}c>{\cellcolor{white}}cccc>{\cellcolor{white}}c>{\cellcolor{white}}c>{\cellcolor{white}}c}
    \Xhline{3\arrayrulewidth}
        \multicolumn{14}{c}{\textbf{\Large{Humor}}}  \\
        \Xhline{3\arrayrulewidth}
        NLD $\rightarrow$ & & \multicolumn{3}{c}{Col} & \multicolumn{3}{c}{POTD} & \multicolumn{3}{c}{HaHa} & \multicolumn{3}{c}{16000} \\  \cmidrule(lr){3-5} \cmidrule(lr){6-8} \cmidrule(lr){9-11} \cmidrule(lr){12-14}
        Model $\downarrow$ & CM & CM+Hi+En & CM+En & CM+Hi & CM+Hi+En & CM+En & CM+Hi & CM+Hi+En & CM+En & CM+Hi & CM+Hi+En & CM+En & CM+Hi \\ \hline \hline
        \rowcolor{gray!10}
        NB & $0.74^{\#}$ & (0.75) & \textbf{(0.75)} & \underline{(0.75)} & \underline{(0.75)} &  (0.75) & (\underline{0.75}) & $0.74^{\#}$ & $0.74^{\#}$ & (0.75) & (\underline{0.75}) & (0.75) & (0.75) \\ 
        \rowcolor{gray!10}
        RF & $0.72^{\#}$ & 0.69 & 0.70 & 0.69 & 0.70 & $0.72^{\#}$ & 0.71 & 0.71 & 0.70 & (0.73) & 0.69 & 0.69 & 0.68 \\ 
        \rowcolor{gray!10}
        SVM & (0.71) & 0.68 & 0.64 & 0.66 & 0.69 & $0.70^{\#}$ & 0.69 & 0.69 & 0.69 & 0.69 & 0.68 & 0.69 & 0.69 \\ 
        \rowcolor{gray!10}
        mBERT & \textbf{(0.78)} & 0.73 & 0.72 & 0.65 & 0.71 & \underline{0.76} & \underline{0.75} & 0.68 & \underline{$0.76^{\#}$} & \underline{$0.76^{\#}$} & 0.74 & 0.75 & 0.70 \\ 
        \rowcolor{gray!10}
        XLM-R & \underline{$0.75^{\#}$} & 0.72 & \underline{0.73} & 0.73 & 0.73 & $0.75^{\#}$ & 0.73 & 0.74 & $0.75^{\#}$ & 0.73 & \textbf{(0.76)} & \underline{(0.76)} & 0.72 \\ 
        \rowcolor{gray!10}
        MuRIL & \underline{0.75} & \textcolor{blue}{($\textbf{0.79}^{\ast}$)} & 0.72 & \textbf{0.76} & 0.73 & \textcolor{blue}{($\textbf{0.79}^{\ast}$)} & $\textbf{0.78}^{\ast \#}$ & \underline{0.77} & \textbf{0.77} & $\textbf{0.78}^{\#}$ & \textbf{0.76} & \textbf{0.77} & $\textbf{0.78}^{\#}$ \\
        \rowcolor{gray!10}
        IndicBERT & 0.74 & \underline{0.76} & 0.72 & 0.71 & $\textbf{0.77}^{\#}$ & \underline{0.76} & \underline{0.75} & \textcolor{blue}{($\textbf{0.79}^{\ast}$)} & \underline{0.76} & 0.75 & 0.74 & 0.73 & \underline{0.76} \\ \hline

    \end{tabular}}

    \vspace{3mm}

    \setkeys{Gin}{keepaspectratio}
    \resizebox*{0.79\textwidth}{0.99\textheight} {
    \begin{tabular}{>{\cellcolor{white}}l|>{\cellcolor{white}}cccc>{\cellcolor{white}}c>{\cellcolor{white}}c>{\cellcolor{white}}cccc}
    \Xhline{3\arrayrulewidth}
        \multicolumn{11}{c}{\textbf{\Large{Sarcasm}}}  \\
        \Xhline{3\arrayrulewidth}
        NLD $\rightarrow$ & & \multicolumn{3}{c}{NHD} & \multicolumn{3}{c}{iSarc} & \multicolumn{3}{c}{SC-V2} \\  \cmidrule(lr){3-5} \cmidrule(lr){6-8} \cmidrule(lr){9-11} 
        Model $\downarrow$ & CM & CM+Hi+En & CM+En & CM+Hi & CM+Hi+En & CM+En & CM+Hi & CM+Hi+En & CM+En & CM+Hi \\ \hline \hline
        \rowcolor{gray!10}
        NB & (0.74) & 0.35 & 0.37 & 0.41 & 0.38 & 0.41 & $0.42^{\#}$ & 0.32 & 0.34 & 0.37 \\ 
        \rowcolor{gray!10}
        RF & (0.69) & 0.43 & 0.51 & $0.60^{\#}$ & 0.51 & 0.57 & 0.57 & 0.49 & $0.60^{\#}$ & 0.58 \\ 
        \rowcolor{gray!10}
        SVM & (0.74) & 0.59 & 0.59 & $0.73^{\#}$ & 0.64 & 0.64 & 0.71 & 0.68 & 0.68 & $0.73^{\#}$ \\ 
        \rowcolor{gray!10}
        mBERT & 0.80 & $\underline{0.83}^{\ast \#}$ & 0.79 & $0.82^{\ast}$ & 0.79 & 0.81 & 0.78 & 0.78 & 0.82 & ($\underline{0.84}^{\ast}$) \\ 
        \rowcolor{gray!10}
        XLM-R & $\underline{0.81}^{\#}$ & ($\underline{0.83}^{\ast}$) & 0.79 & ($\underline{0.83}^{\ast}$) & \underline{$0.81^{\#}$} & $0.81^{\#}$ & $0.81^{\#}$ & 0.80 & (\underline{0.83}) & $0.81^{\#}$ \\ 
        \rowcolor{gray!10}
        MuRIL & \textbf{0.83} & $\textbf{0.86}^{\ast}$ & \textcolor{blue}{($\textbf{0.89}^{\ast}$)} & 0.82 & \textbf{0.84} & $\underline{0.85}^{\ast}$ & \textbf{0.84} & $\textbf{0.87}^{\ast \#}$ & \textbf{0.84} & $\textbf{0.87}^{\ast \#}$ \\ 
        \rowcolor{gray!10}
        IndicBERT & \underline{0.81} & $\textbf{0.86}^{\ast \#}$ & $\underline{0.86}^{\ast \#}$ & $\textbf{0.85}^{\ast}$ & \underline{0.81} & ($\textbf{0.88}^{\ast}$) & \underline{0.83} & \underline{0.83} & \underline{0.83} & 0.82 \\ \hline

    \end{tabular}}

    \caption{Results of our experiment evaluating the impact of mixing native samples for humor (upper table) and sarcasm (lower table) detection. Notation: NLD for native language dataset, CM for code-mixed. Reported scores are $F1$ scores of the positive class and are averaged over three different random seeds.}
    \label{tab:results_native_humor}
\end{table*}
\endgroup

In this section, we reported our observations from the first experiment. The F1-scores obtained from all models over the considered datasets and training scenarios are reported in Table \ref{tab:results_native_humor}. The F1-scores of best-performing models for the individual training scenarios (column-wise in Table \ref{tab:results_native_humor}) were marked in bold, while the second-best results are underlined. Similarly, the best-performing scenarios giving the highest $F1$ scores for individual models (row-wise in Table \ref{tab:results_native_humor}) were kept inside the parenthesis, while the second best scores are marked with `$\#$' superscript. The highest score for both tasks across all training scenarios and models are marked in \textcolor{blue}{blue}. Following are our takeaways,

\begin{itemize}

    \item When trained with only code-mixed samples, mBERT gave the highest F1-score of \textbf{0.78} for humour detection, followed by XLM-R and MuRIL (F1-score of \textbf{0.75}). Similarly, for sarcasm detection, MuRIL reported the highest F1-score of \textbf{0.83} followed by IndicBERT and XLM-R (F1-score of \textbf{0.81}).  
    
    \item On adding native samples to the code-mixed training sets, statistical classifiers did not show any performance improvement. In fact, in many cases, the performance declined sharply (a decline of \textbf{9.85\%} and \textbf{56.7\%} in F1-scores for humour and sarcasm detection, respectively). 
    
    

    \item Among the MLMs, mBERT showed significant improvement after native sample mixing with an F1-score of \textbf{0.84} (improvements up to \textbf{5\%}) in detecting sarcasm. In the case of MuRIL, we saw many instances where it resulted in statistically significant (p < 0.05) improvement (up to \textbf{5.3\%} and \textbf{7.2\%} raise in F1-score for humour and sarcasm detection respectively) after native sample mixing. Finally, we observed that IndicBERT model performed best after native sample mixing with an F1 scores of \textbf{0.79} (improvement up to \textbf{6.7\%}) and \textbf{0.88} (improvement up to \textbf{8.6\%}) in code-mixed humour and sarcasm detection respectively (both statistically significant with p< 0.05).
    
    

    

    \item We observed that MuRIL and IndicBERT gave the overall highest scores. This is interesting as both are special LLMs exclusively developed for Indian languages. This observation is consistent with \citet{mazumder2024improving} as the same phenomenon was observed for code-mixed hate detection. Another important observation was that the addition of Hindi samples didn't result in a sharp improvement in the F1-score as we saw in \citet{mazumder2024improving}. On inspection, we found that the synthetic samples are not humorous and sarcastic compared to their original English ones. In other words, the humour and sarcasm got lost during translation. A detailed discussion of the same is reported in Appendix \ref{appendix:translation_hin}.

    

    \item We observed similar trends in precision and recall scores presented in Appendix Table \ref{tab:results_native_humor_precall}.
\end{itemize}

\subsection{Observations from Exp. 2:}
\label{Exp. 2 Results}

\afterpage{
\begingroup
\renewcommand{\arraystretch}{1.08} 
\begin{table}
    \centering
    \setkeys{Gin}{keepaspectratio}
    \resizebox*{\columnwidth}{0.99\textheight} {
    \begin{tabular}{ll|cccccc}
    \Xhline{3\arrayrulewidth}
        \multicolumn{8}{c}{\textbf{\Large{Humor}}}  \\
        \Xhline{3\arrayrulewidth}
        \multicolumn{2}{c|}{NLD : \textit{Col}} &  \multicolumn{2}{c}{$\text{mBERT}_{MTL}$}  & \multicolumn{2}{c}{$\text{XLM-R}_{MTL}$} & \multicolumn{2}{c}{$\text{MuRIL}_{MTL}$}  \\  \cmidrule(lr){1-2} \cmidrule(lr){3-4} \cmidrule(lr){5-6} \cmidrule(lr){7-8}
        Hate & Sarcasm  & Gate & w/o Gate & Gate & w/o Gate & Gate & w/o Gate \\ \hline \hline
        \checkboxempty & $\checkboxcmark_{NHD}$ & 0.69 & 0.76 & $\underline{0.78}^{\ast}$ & $\textbf{0.79}^{\ast \#}$ & 0.76 & 0.68 \\
        \checkboxempty & $\checkboxcmark_{iSarc}$ & 0.71 & \textbf{0.78} & \underline{0.76} & 0.75 & \underline{0.76} & 0.71 \\
        \checkboxempty & $\checkboxcmark_{SC-V2}$ & 0.67 & \underline{0.76} & $\textbf{0.80}^{\ast \#}$ & ($\textbf{0.80}^{\ast}$) & 0.74 & \underline{0.76} \\
        $\checkboxcmark$ & \checkboxempty & \textbf{0.77} & 0.72 & 0.71 & \underline{0.75} & \textbf{0.77} & \underline{0.75} \\
        $\checkboxcmark$ & $\checkboxcmark_{NHD}$ & \underline{0.78} & $\textbf{0.79}^{\#}$ & $\textbf{0.79}^{\ast}$ & $\textbf{0.79}^{\ast \#}$ & 0.76 & $\underline{0.78}^{\#}$ \\
        $\checkboxcmark$ & $\checkboxcmark_{iSarc}$ & \underline{0.78} & 0.77 & $\textbf{0.79}^{\ast}$ & \underline{0.78} & $\textbf{0.79}^{\ast}$ & $0.76^{\ast}$ \\
        $\checkboxcmark$ & $\checkboxcmark_{SC-V2}$ & \underline{0.78} & $\textbf{0.79}^{\#}$ & $\textbf{0.79}^{\ast}$ & $\underline{0.78}^{\ast}$ & 0.75 & 0.75 \\ \hline
        
        \multicolumn{2}{c|}{NLD : \textit{POTD}}  & & & & & \\ \cmidrule(lr){1-2}
        \checkboxempty & $\checkboxcmark_{NHD}$ & 0.71 & $0.79^{\#}$ & $\underline{0.79}^{\ast}$ & ($\textbf{0.80}^{\ast}$) & 0.73 & 0.71 \\
        \checkboxempty & $\checkboxcmark_{iSarc}$ & 0.76 & $\textbf{0.79}^{\#}$ & 0.78 & $\underline{0.78}^{\ast}$ & 0.76 & $\underline{0.78}^{\ast \#}$ \\
        \checkboxempty & $\checkboxcmark_{SC-V2}$ & 0.71 & \underline{$0.79^{\#}$} & 0.70 & ($\textbf{0.80}^{\ast}$) & 0.75 & 0.72 \\
        $\checkboxcmark$ & \checkboxempty & 0.69 & $\textbf{0.79}^{\#}$ & \underline{0.75} & $\textbf{0.79}^{\ast \#}$ & 0.67 & 0.70 \\
        $\checkboxcmark$ & $\checkboxcmark_{NHD}$ & \underline{$0.79^{\#}$} & $\underline{0.79}^{\#}$ & $\textbf{0.80}^{\ast \#}$ & $\underline{0.79}^{\ast \#}$ & 0.74 & 0.75 \\
        $\checkboxcmark$ & $\checkboxcmark_{iSarc}$ & $\textbf{0.79}^{\#}$ & \underline{0.78} & $\textbf{0.79}^{\ast}$ & $\underline{0.78}^{\ast}$ & \underline{0.78} & $\underline{0.78}^{\ast \#}$ \\
        $\checkboxcmark$ & $\checkboxcmark_{SC-V2}$ & \underline{$0.79^{\#}$} & $\underline{0.79}^{\#}$ & ($\textbf{0.81}^{\ast}$) & $0.71^{\ast}$ & 0.69 & 0.75 \\ \hline
        
        \multicolumn{2}{c|}{NLD : \textit{HaHa}}  & & & & & \\ \cmidrule(lr){1-2}
        \checkboxempty & $\checkboxcmark_{NHD}$ & 0.77 & 0.78 & $\textbf{0.80}^{\ast \#}$ & $\underline{0.78}^{\ast}$ & 0.71 & 0.76 \\
        \checkboxempty & $\checkboxcmark_{iSarc}$ & 0.76 & \underline{0.78} & $\textbf{0.80}^{\ast \#}$ & 0.75 & 0.69 & 0.75 \\
        \checkboxempty & $\checkboxcmark_{SC-V2}$ & $\textbf{0.79}^{\#}$ & \underline{0.78} & \underline{0.78} & $\underline{0.78}^{\ast}$ & 0.71 & 0.75 \\
        $\checkboxcmark$ & \checkboxempty & 0.72 & 0.71 & \underline{0.77} & 0.76 & $\textbf{0.80}^{\ast \#}$ & 0.76 \\
        $\checkboxcmark$ & $\checkboxcmark_{NHD}$ & 0.77 & \underline{$0.79^{\#}$} & $\textbf{0.80}^{\ast \#}$ & $\underline{0.79}^{\ast \#}$ & $\underline{0.79}^{\ast}$ & 0.75 \\
        $\checkboxcmark$ & $\checkboxcmark_{iSarc}$ & \underline{0.78} & ($\textbf{0.80}^{\ast}$) & 0.77 & 0.76 & \underline{0.78} & $\underline{0.78}^{\#}$ \\
        $\checkboxcmark$ & $\checkboxcmark_{SC-V2}$ & (0.80) & $0.79^{\#}$ & $0.78^{\ast}$ & 0.78 & \textcolor{blue}{($\textbf{0.83}^{\ast}$)} & ($\underline{0.81}^{\ast}$) \\ \hline
        
        \multicolumn{2}{c|}{NLD : \textit{16000}}  & & & & & \\ \cmidrule(lr){1-2}
        \checkboxempty & $\checkboxcmark_{NHD}$ & 0.76 & 0.68 & $\underline{0.79}^{\ast}$ & ($\textbf{0.80}^{\ast}$) & 0.69 & 0.76 \\
        \checkboxempty & $\checkboxcmark_{iSarc}$ & 0.76 & \underline{0.78} & $\textbf{0.80}^{\ast \#}$ & 0.77 & $\underline{0.78}^{\ast}$ & 0.69 \\
        \checkboxempty & $\checkboxcmark_{SC-V2}$ & $\textbf{0.79}^{\#}$ & 0.68 & \underline{0.78} & $\textbf{0.79}^{\ast \#}$ & 0.73 & 0.68 \\
        $\checkboxcmark$ & \checkboxempty & \textbf{0.78} & \textbf{0.78} & \underline{0.77} & $\textbf{0.78}^{\ast}$ & \underline{0.77} & 0.70 \\
        $\checkboxcmark$ & $\checkboxcmark_{NHD}$ & 0.76 & $\textbf{0.79}^{\#}$ & 0.77 & $\textbf{0.79}^{\ast \#}$ & \underline{0.78} & 0.76 \\
        $\checkboxcmark$ & $\checkboxcmark_{iSarc}$ & \textbf{0.78} & \underline{0.77} & \underline{0.77} & 0.75 & \textbf{0.78} & 0.76 \\
        $\checkboxcmark$ & $\checkboxcmark_{SC-V2}$ & \underline{0.78} & ($\textbf{0.80}^{\ast}$) & $\textbf{0.80}^{\ast \#}$ & 0.77 & $\textbf{0.80}^{\ast \#}$ & 0.77 \\ \hline
    \end{tabular}}

    \vspace{5mm}

    \setkeys{Gin}{keepaspectratio}
    \resizebox*{\columnwidth}{0.99\textheight} {
    \begin{tabular}{ll|cccccc}
    \Xhline{3\arrayrulewidth}
        \multicolumn{8}{c}{\textbf{\Large{Sarcasm}}}  \\
        \Xhline{3\arrayrulewidth}
        \multicolumn{2}{c|}{NLD : \textit{NHD}} &  \multicolumn{2}{c}{$\text{mBERT}_{MTL}$}  & \multicolumn{2}{c}{$\text{XLM-R}_{MTL}$} & \multicolumn{2}{c}{$\text{MuRIL}_{MTL}$}  \\  \cmidrule(lr){1-2} \cmidrule(lr){3-4} \cmidrule(lr){5-6} \cmidrule(lr){7-8}
        Hate & Humor  & Gate & w/o Gate & Gate & w/o Gate & Gate & w/o Gate \\ \hline \hline
        \checkboxempty & $\checkboxcmark_{Col}$ & 0.82 & \underline{0.83} & $\textbf{0.85}^{\ast}$ & 0.82 & \underline{0.83} & 0.78  \\
        \checkboxempty & $\checkboxcmark_{POTD}$ & \underline{0.84} & 0.83 & $\textbf{0.85}^{\ast}$ & 0.82 & 0.82 & 0.76 \\
        \checkboxempty & $\checkboxcmark_{HaHa}$ & \underline{0.84} & 0.83 & $\textbf{0.85}^{\ast}$ & 0.82 & \underline{0.84} & 0.78 \\
        \checkboxempty & $\checkboxcmark_{16000}$ & 0.81 & 0.83 & $\textbf{0.90}^{\ast \#}$ & 0.82 & $\underline{0.86}^{\ast}$ & 0.79 \\
        \checkboxcmark & $\checkboxempty$ & 0.81 & $\underline{0.85}^{\ast \#}$ & $\textbf{0.86}^{\ast}$ & 0.82 & \textbf{0.86} & 0.78 \\
        \checkboxcmark & $\checkboxcmark_{Col}$ & $0.83^{\ast}$ & ($\underline{0.86}^{\ast}$) & $0.84^{\ast}$ & $\textbf{0.88}^{\ast \#}$ & $\underline{0.86}^{\ast}$ & (0.84) \\
        \checkboxcmark & $\checkboxcmark_{POTD}$ & $0.84^{\ast}$ & 0.81 & $\textbf{0.88}^{\ast}$ & $\underline{0.87}^{\ast}$ & 0.84 & 0.82 \\
        \checkboxcmark & $\checkboxcmark_{HaHa}$ & $0.84^{\ast}$ & $0.83^{\ast}$ & $\textbf{0.88}^{\ast}$ & $\underline{0.87}^{\ast}$ & 0.81 & 0.81 \\
        \checkboxcmark & $\checkboxcmark_{16000}$ & $0.85^{\ast \#}$ & $0.85^{\ast \#}$ & $\textbf{0.88}^{\ast}$ & 0.82 & $\underline{0.87}^{\ast \#}$ & 0.81 \\ \hline
        
        \multicolumn{2}{c|}{NLD : \textit{iSarc}}  & & & & & \\ \cmidrule(lr){1-2}
        \checkboxempty & $\checkboxcmark_{Col}$ & 0.81 & \underline{0.83} & $\underline{0.83}^{\ast}$ & 0.81 & \textbf{0.86} & 0.78 \\
        \checkboxempty & $\checkboxcmark_{POTD}$ & $\underline{0.83}^{\ast}$ & $\textbf{0.85}^{\ast \#}$ & \underline{0.83} & 0.80 & 0.82 & 0.79 \\
        \checkboxempty & $\checkboxcmark_{HaHa}$ & $0.84^{\ast}$ & 0.81 & $\underline{0.85}^{\ast}$ & 0.81 & \textbf{0.86} & 0.79 \\
        \checkboxempty & $\checkboxcmark_{16000}$ & 0.81 & $\textbf{0.85}^{\ast \#}$ & \underline{0.82} & 0.81 & \underline{0.82} & 0.78 \\
        \checkboxcmark & $\checkboxempty$ & 0.82 & $\underline{0.83}^{\ast}$ & 0.82 & 0.81 & \textbf{0.85} & 0.72 \\
        \checkboxcmark & $\checkboxcmark_{Col}$ & 0.81 & 0.80 & $\underline{0.86}^{\ast}$ & 0.82 & ($\textbf{0.88}^{\ast}$) & 0.79 \\
        \checkboxcmark & $\checkboxcmark_{POTD}$ & 0.82 & 0.74 & $\textbf{0.88}^{\ast}$ & \underline{0.83} & 0.82 & 0.79 \\
        \checkboxcmark & $\checkboxcmark_{HaHa}$ & 0.81 & 0.81 & $\textbf{0.88}^{\ast}$ & 0.82 & \underline{0.85} & 0.80 \\
        \checkboxcmark & $\checkboxcmark_{16000}$ & 0.81 & 0.81 & $\textbf{0.88}^{\ast}$ & $\underline{0.85}^{\ast}$ & 0.82 & 0.79 \\ \hline

        \multicolumn{2}{c|}{NLD : \textit{SC-V2}}  & & & & & \\ \cmidrule(lr){1-2}
        \checkboxempty & $\checkboxcmark_{Col}$ & $\textbf{0.84}^{\ast}$ & $\textbf{0.84}^{\ast}$ & \underline{0.83} & \underline{0.83} & 0.81 & $\underline{0.83}^{\#}$ \\
        \checkboxempty & $\checkboxcmark_{POTD}$ & 0.82 & 0.82 & $\underline{0.84}^{\ast}$ & 0.83 & \textbf{0.85} & $0.83^{\#}$ \\
        \checkboxempty & $\checkboxcmark_{HaHa}$ & $\textbf{0.85}^{\ast \#}$ & $\underline{0.83}^{\ast}$ & 0.82 & \underline{0.83} & \textbf{0.85} & $\underline{0.83}^{\#}$ \\
        \checkboxempty & $\checkboxcmark_{16000}$ & 0.81 & $\underline{0.84}^{\ast}$ & $\textbf{0.85}^{\ast}$ & 0.83 & \underline{0.84} & $0.83^{\#}$ \\
        \checkboxcmark & $\checkboxempty$ & $0.83^{\ast}$ & $0.84^{\ast}$ & $\textbf{0.88}^{\ast}$ & 0.83 & \underline{0.85} & 0.79 \\
        \checkboxcmark & $\checkboxcmark_{Col}$ & 0.80 & 0.82 & $\textbf{0.89}^{\ast}$ & $\underline{0.87}^{\ast}$ & 0.84 & 0.81 \\
        \checkboxcmark & $\checkboxcmark_{POTD}$ & $0.83^{\ast}$ & $0.83^{\ast}$ & $\textbf{0.89}^{\ast}$ & $0.85^{\ast}$ & $\underline{0.86}^{\ast}$ & (0.84) \\
        \checkboxcmark & $\checkboxcmark_{HaHa}$ & ($0.86^{\ast}$) & $0.84^{\ast}$ & $0.85^{\ast}$ & ($\textbf{0.89}^{\ast}$) & $\underline{0.87}^{\ast \#}$ & 0.79 \\
        \checkboxcmark & $\checkboxcmark_{16000}$ & $0.83^{\ast}$ & 0.81 & \textcolor{blue}{($\textbf{0.91}^{\ast}$)} & $\underline{0.86}^{\ast}$ & 0.85 & 0.81 \\ \hline
    \end{tabular}}

    \caption{Results of our experiment of multi-task learning and ablation study for humor (upper table) and sarcasm (lower table) detection. Notation: `NLD' for Native Language Dataset. Reported scores are $F1$ scores of the positive class and are averaged over three different random seeds.}
    \label{tab:MTL_humor_results}
\end{table}
\endgroup
}

In this section, we presented our observations from the second experiment. The F1-scores obtained from all models for the considered datasets and training scenarios are reported in Table \ref{tab:MTL_humor_results}. In Table \ref{tab:MTL_humor_results}, each row enlists different training scenarios for considered two tasks. It has two sub-tables. The upper and lower tables report results for code-mixed humour and sarcasm detection, respectively. The first two columns in each sub-table report the combination of datasets used for training. For example, the first row under the ‘Humor’ sub-table presents the case where the training set has (i) humour samples from the code-mixed dataset and English dataset ‘ColBERT’ (written as NLD: ‘Col’), (ii) sarcasm samples from the code-mixed dataset and English dataset ‘NHD’ (tick marked next to NHD). We didn’t consider hate samples here (presented as the empty box under the ‘Hate’ column). So, we trained our models for two tasks in this case. Similarly, the fifth row represents the case where the training set has code-mixed, native English and native Hindi hate samples along with the Humour and Sarcasm samples considered in the first row. So, in this case, we trained our models for three tasks.
The overall best scores were marked in \textcolor{blue}{blue}. Following were our takeaways,

\begin{itemize}
    \item For code-mixed humor detection, $\text{MuRIL}_{MTL}$ reported the highest F1 score of \textbf{0.83} (up to \textbf{10.67\%} increment), followed by $\text{XLM-R}_{MTL}$ (\textbf{0.81}) and $\text{mBERT}_{MTL}$ (\textbf{0.80}). On the other hand, for code-mixed sarcasm detection, $\text{XLM-R}_{MTL}$ outperformed others with \textbf{0.91} F1 score (up to \textbf{12.35\%} increment), followed by $\text{MuRIL}_{MTL}$ (\textbf{0.88}) and $\text{mBERT}_{MTL}$ (\textbf{0.86}).

    \item We achieved the highest scores in MTL strategy when native datasets with \textit{low KL divergence} between classes were combined. This appears to help the pre-trained MLM-based MTL architecture focus more on contextual understanding rather than being influenced by lexical differences between the labels. Notably, SC-V2 and HCHIn, which have the lowest KL divergence, were consistently present among the best-performing configurations in both sub-tables of Table \ref{tab:MTL_humor_results} (refer [row 26, col 7] of Humor subtable and [row 32, col 5] of Sarcasm subtable).

    \item For sarcasm detection, $\text{mBERT}_{MTL}$ resulted with a highest F1 score of \textbf{0.86} (improvement up to \textbf{7.5\%}). The improvement is statistically significant ($p<0.05$).

    \item Similarly, for humor detection, $\text{XLM-R}_{MTL}$ gave an improvement of F1 score upto \textbf{0.81} (improvement up to \textbf{8\%}). For sarcasm detection, $\text{XLM-R}_{MTL}$ improved even more with the highest F1 score of \textbf{0.91} (improvement up to \textbf{12.35\%}). Both the improvements were statistically significant with $p<0.05$.

    \item Finally, $\text{MuRIL}_{MTL}$ reported the highest F1 score for humor detection, i.e., \textbf{0.83} (an improvement up to \textbf{10.67\%}). The improvement is statistically significant ($p<0.05$). Similarly, for sarcasm detection, F1 scores ranged from \textbf{0.81} to \textbf{0.88} using $\text{MuRIL}_{MTL}$. The F1-score improved up to \textbf{6\%} which is statistically significant ($p<0.05$).


    \item Upon analyzing the improvements deeply, we found that the MTL models performed better for samples with \textit{shorter context} length (refer to Figure \ref{fig:context_len} in Appendix \ref{section:Ablation Study}) and \textit{spelling errors}. Due to space constraints, we reported these insights from the ablation study in Appendix \ref{section:Ablation Study}.

    \item The error analysis (refer Section \ref{section:Error Analysis}) revealed that samples with some connection to the hate detection task improved performance in multitask setting.

    \item Similar patterns appeared in the precision and recall scores (Table \ref{tab:MTL_humor_results_precall} and Table \ref{tab:MTL_sarcasm_results_precall}) as reported in the Appendix.

    

\end{itemize}

\subsection{Observations from Exp. 3:}
\label{Exp. 3 Results}

\begingroup
\renewcommand{\arraystretch}{1.08} 
\begin{table}
    \centering
    \setkeys{Gin}{keepaspectratio}
    \resizebox*{\columnwidth}{0.14\textwidth} {
    \begin{tabular}{l|ccccc}
    \Xhline{3\arrayrulewidth}
        \multicolumn{6}{c}{\textbf{\Large{Humor}}}  \\
        \Xhline{3\arrayrulewidth}
        Model & CM & Col & POTD & HaHa & 16000 \\ \hline \hline
        Gemma & 0.09 & 0.09 & 0.07 & (0.11) & $0.10^{\#}$ \\
        Aya Expanse & 0.73 & (\textbf{0.74}) & $\underline{0.73}^{\#}$ & \underline{0.71} & \underline{0.73}    \\
        Llama-3.1 & \textcolor{blue}{(\textbf{0.75})} & $\textbf{0.74}^{\#}$ & \textcolor{blue}{\textbf{(0.75)}} & \textcolor{blue}{\textbf{(0.75)}} & \textcolor{blue}{\textbf{(0.75)}} \\ 
        GPT-4 & (\underline{0.74}) & \underline{0.57} & $0.62^{\#}$ & 0.56 & 0.58 \\ \hline

    \end{tabular}
    }

    \vspace{3mm}

    \setkeys{Gin}{keepaspectratio}
    \resizebox*{\columnwidth}{0.125\textwidth} {
    \begin{tabular}{l|cccc}
    \Xhline{3\arrayrulewidth}
        \multicolumn{5}{c}{\textbf{\Large{Sarcasm}}}  \\
        \Xhline{3\arrayrulewidth}
        Model & CM & NHD & iSarc & SC-V2 \\ \hline \hline
        Gemma & \underline{0.34} & 0.11 & \underline{$0.39^{\#}$} & (\textbf{0.47}) \\
        Aya Expanse &  0.21 & \underline{0.21} & (0.26) & $0.23^{\#}$ \\
        Llama-3.1 & 0.21 & $\textbf{0.45}^{\#}$ & \textbf{(0.51)} & 0.28 \\ 
        GPT-4 & \textcolor{blue}{(\textbf{0.78})} & 0.17 & 0.25 & \underline{$0.43^{\#}$}  \\ \hline

    \end{tabular}
    }

    \caption{Results of our experiment evaluating the impact of in-context learning on VMLMs. Notation: CM for code-mixed.}
    \label{tab:results_vmlm}
\end{table}
\endgroup
\begingroup
\renewcommand{\arraystretch}{1.08} 
\begin{table}
    \centering
    \setkeys{Gin}{keepaspectratio}
    \resizebox*{\columnwidth}{0.14\textwidth} {
    \begin{tabular}{l|ccccc}
    \Xhline{3\arrayrulewidth}
        \multicolumn{6}{c}{\textbf{\Large{Humor}}}  \\
        \Xhline{3\arrayrulewidth}
        Model & CM & CM+Col & CM+POTD & CM+HaHa & CM+16000 \\ \hline \hline
        Gemma & \textcolor{blue}{\textbf{(0.77)}} & 0.58 & \textcolor{blue}{\textbf{(0.77)}} & 0.23 & \textbf{0.76} \\
        Aya Expanse & 0.74 & \textbf{0.74} & 0.75 & \textbf{0.74} & \textbf{(0.76)}    \\
        Llama-3.1 & \textcolor{blue}{\textbf{(0.77)}} & \textbf{0.74} & 0.75 & \textbf{0.74} & 0.75 \\ \hline

    \end{tabular}
    }

    \vspace{3mm}

    \setkeys{Gin}{keepaspectratio}
    \resizebox*{\columnwidth}{0.125\textwidth} {
    \begin{tabular}{l|cccc}
    \Xhline{3\arrayrulewidth}
        \multicolumn{5}{c}{\textbf{\Large{Sarcasm}}}  \\
        \Xhline{3\arrayrulewidth}
        Model & CM & CM+NHD & CM+iSarc & CM+SC-V2 \\ \hline \hline
        Gemma & (0.74) & 0.63 & 0.67 & 0.64 \\
        Aya Expanse &  0.78 & \textbf{0.68} & \textbf{(0.79)} & 0.78 \\
        Llama-3.1 & \textbf{0.80} & 0.49 & 0.70 & \textcolor{blue}{\textbf{(0.81)}} \\ \hline

    \end{tabular}
    }

    \caption{Results of our experiment evaluating the impact of native language mixing in instruction fine-tuning of VMLMs using LoRA adapter. Notation: CM for code-mixed.}
    \label{tab:results_vmlm_lora}
\end{table}
\endgroup

In this section, we reported our observations from the third experiment. The best F1-scores obtained from prompting VMLMs are reported in Table \ref{tab:results_vmlm}. A more detailed overview of obtained F1-scores is presented in Table \ref{tab:results_vmlm_all_fewshot}. The F1-scores obtained from instruction finetuning VMLMs are presented in Table \ref{tab:results_vmlm_lora}.
The highest overall scores for both tasks are marked in \textcolor{blue}{blue}.
Following were our takeaways,

\begin{itemize}
    \item When we prompted the VMLMs with code-mixed few shots, Llama-3.1 achieved the highest F1-score (\textbf{0.75}) for humour detection, followed by GPT-4 (\textbf{0.74}) and Aya Expanse (\textbf{0.73}). In contrast, Gemma performed the worst with an F1-score of \textbf{0.09}. For sarcasm detection, GPT-4 outperformed all models with an F1-score of \textbf{0.78}, while others showed a sharp decline, i.e., Gemma (\textbf{0.34}), Aya Expanse (\textbf{0.21}), and Llama-3.1 (\textbf{0.21}).

    \item When prompted with native humour few-shot examples, Llama-3.1 maintained a stable F1-score between \textbf{0.74–0.75}, while Aya Expanse showed no significant improvement, maintaining scores in the \textbf{0.71–0.74} range. In contrast, Gemma performed poorly, with F1-scores ranging from \textbf{0.07–0.11}. GPT-4 also experienced a decline in performance, with F1-scores dropping to \textbf{0.56–0.62}. 
    Native few-shot prompting led to some improvements in sarcasm detection across models. Gemma’s performance significantly increased to \textbf{0.47}, compared to \textbf{0.34} with code-mixed few-shots. Similarly, Aya Expanse improved to \textbf{0.26}, up from \textbf{0.21}, while Llama-3.1 achieved an F1-score of \textbf{0.51}, a substantial increase from \textbf{0.21} in the code-mixed setting. However, GPT-4 continued to struggle, with F1-scores ranging from \textbf{0.17–0.43}.

    \item When VMLMs were finetuned using the code-mixed training set, Gemma and Llama-3.1 achieved the best F1-score of \textbf{0.77} in humour detection, while Llama-3.1 got best F1-score of \textbf{0.80} in sarcasm detection. When finetuned using native sample mixing strategy, Aya Expanse and Llama-3.1 maintained comparable performances within the range \textbf{0.74-0.76}, whereas Gemma showed greater fluctuations (\textbf{0.23-0.77}). For sarcasm detection also, VMLMs showed varying fluctuations, i.e., Gemma (\textbf{0.63-0.67}), Aya Expanse (\textbf{0.68-0.79}), and Llama-3.1 (\textbf{0.49-0.81}). Although this finetuning approach outperformed the few-shot prompting method, it did not exceed the best F1-scores obtained with MLMs. Futhermore, a consistent pattern emerged from the native dataset point of view, where \textit{low KL divergence} native datasets proved to be more effective for training compared to others.

    \item The precision and recall scores showed a similar pattern. The exact values are reported in Appendix Table \ref{tab:results_vmlm_precall} and Table \ref{tab:results_vmlm_lora_precall}.

\end{itemize}

\section{Error analysis:} 
\label{section:Error Analysis}

\afterpage{
\begingroup
\renewcommand{\arraystretch}{1.2} 
\begin{table*}[ht]
    \centering
    \setkeys{Gin}{keepaspectratio}
    \resizebox*{1.0\textwidth}{1.0\textheight} {
    \begin{tabular}{|p{0.1in}|p{2.1in}|p{2.1in}|ccc|ccc|ccc|}
    \hline
        \multirow{2}{0.7em}{Sl No} & \multirow{2}{12em}{Sample} & \multirow{2}{10em}{Translated English} & \multicolumn{3}{c|}{CM} & \multicolumn{3}{c|}{NSM} & \multicolumn{3}{c|}{MTL} \\ \cline{4-12}
         &  & & M1 & M2 & M3 & M1 & M2 & M3 &  M1 & M2 & M3 \\   \hline \hline
        1 & .@shashitharoor sir Kejriwal power cut nahi karenge to bill kam kaise hoga? \textbf{(Humor)} & .@shashitharoor sir Kejriwal power cut nahi karenge to bill kam kaise hoga? & $\times$ & $\times$ & $\times$ & $\times$ & $\times$ & $\times$ & $\times$  & \checkmark & \checkmark \\   \hline
        2 & Musalmaano ka intolerance kuch zyada hi badh raha hai.. Par Media gaalia sirf Hindu ko deti hai \textbf{(Non-humor)} & The intolerance among Muslims seems to be increasing excessively... But the media abuses only Hindus. & \checkmark & \checkmark & \checkmark & $\times$ & \checkmark  & $\times$  & $\times$  & $\times$ & $\times$ \\   \hline
        3 & Kehte hain Agar kisi cheez ko dil se chaaho to puri kayanat usey tumse milane ki koshish mein lag jaati hai. \#dada is back \#ipl4 \#srk \#irony \textbf{(Non-sarcasm)} & They say if you truly desire something from the heart, the whole universe conspires to make it happen. \#dada is back \#ipl4 \#srk \#irony & \checkmark & \checkmark & \checkmark & $\times$ & $\times$ & \checkmark & $\times$ & $\times$ & $\times$ \\   \hline
        4 & Culturally rich honge..par gavaaro ki basti bhi mera desh. \#SuSaid \#irony \#india \textbf{(Sarcasm)} & They might be culturally rich, but my country is also a haven for the uneducated. \#SuSaid \#irony \#india & $\times$ & $\times$ & $\times$ & \checkmark & $\times$ & $\times$ & \checkmark  & \checkmark  & $\times$ \\   \hline
        5 & @RahulBose1 Fir bhi mera bharat maahan. \#Sarcasm \textbf{(Non-sarcasm)} & @RahulBose1 Yet, my India is still great. \#Sarcasm & \checkmark & \checkmark & \checkmark & $\times$ & $\times$ & $\times$ & $\times$ & \checkmark & $\times$ \\   \hline

    \end{tabular}}

    \caption{Selected examples for various cases reported under error analysis. Here, the `\checkmark', and the `$\times$' denote correct and incorrect classification by the corresponding model, respectively. Notation: CM for code-mixed, NSM for native sample mixing, MTL for Multi-Task Learning; M1 for mBERT, M2 for XLM-R and M3 for MuRIL. The columns under \textbf{CM} reported the results when the models were trained with only code-mixed samples and the columns under \textbf{MTL} reported the results of the best performing MTL model for each task.}
    \label{tab:example_ablation}
\end{table*}
\endgroup
}

To better understand the models' errors, we conducted a qualitative error analysis by examining some correctly and incorrectly classified samples, as presented in Table \ref{tab:example_ablation}. We observed the following:

\begin{itemize}
    \item For the ironic humor in \textit{Sl. No. 1}, 
    the humor arises from the switch between the political promise (\textit{'no power cuts'}) and the ironic consequence (\textit{'high bills'}) of the situation. Most of the models failed on it, except $\text{XLM-R}_{MTL}$ and $\text{MuRIL}_{MTL}$ as it had source of knowledge from other tasks like sarcasm.
    \item \textit{Sl. No. 2} shows how all the MTL models struggled with non-humorous sample related to religious domain containing keywords like \textit{`intolerance'} and \textit{`gaalia'}, likely due to task interference from the hate detection task, where these keywords are often used in hateful contexts.
    \item In \textit{Sl. No. 3} and \textit{4}, despite keywords like \textit{\#Sarcasm}' and \textit{\#irony}', models trained on code-mixed data accurately predicted non-sarcastic contexts, whereas NSM and MTL models failed.
    \item MTL models effectively captured sarcasm in hateful contexts by combining hate detection with other tasks. For example, in \textit{Sl. No. 5}, the word \textit{`gavaaro'} (\textbf{Gloss:} uneducated) conveys explicit hate and MTL models identified the sarcastic tone in it.
\end{itemize}

\section{Conclusion:}

From our findings, we drew the following conclusions: 

\begin{itemize}

    \item Among these three strategies, MTL reported the most significant improvement, with F1-score increments upto \textbf{10.67\%} for humor and \textbf{12.35\%} for sarcasm. Native sample mixing followed, with increments upto \textbf{6.76\%} for humor and \textbf{8.64\%} for sarcasm, while VMLMs in both set-up showed no improvement in F1-scores.

    \item  The ablation study highlighted the importance of the gating mechanism within the MTL framework, particularly in samples with \textit{`shorter context lengths'} and those containing \textit{`misspelled words'}.

    \item The error analysis (refer Section \ref{section:Error Analysis}) presented samples which justified the utility of related tasks in the multitask scenario.

    \item The VMLMs showed poor performance due to a tendency to favor specific labels in sentiment-related classification tasks (refer Appendix \ref{Detailed Exp. 3 Results} for detailed observations). This insight aligned well with previous works \cite{Ges2023IsGG, baranov-etal-2023-told, zhang2024sarcasmbench} which showed LLMs inability in detecting monolingual humor and sarcasm.
    



\end{itemize}

\section{Limitation:}

In this section, we reported some of the limitations of our work.

\begin{itemize}
    \item Our MTL models occasionally failed on non-humorous and non-sarcastic samples. The possible reason behind it could be task and domain interference. We reported a detailed qualitative analysis of correctly and incorrectly classified samples in error analysis section (refer Appendix \ref{section:Error Analysis}).  
    
    \item Non-availability of native Hindi language datasets for these two tasks restricted us from utilizing gold labeled Hindi samples for mixing in our experiments. 
    
    \item Due to the limited linguistic expertise, we evaluated our hypothesis only on the Hindi-English code-mixed scenarios. Other language pairs can be utilized to shed some light on the generalization of our approach to more languages.
    
    \item We couldn't make use of more larger VMLMs due to computational constraints. The larger ones (with more than 100B parameters) may generalize the results more clearly.
    
    \item Here, in our experiments we utilized the most widely used translator (Google Translate API) to generate synthetic Hindi samples. One can try other possible ways for synthetic data generation using various VMLMs, in the same lines as future scope.
    
    \item We explored the impact of native samples for code-mixed humor and sarcasm detection. As future scope, one can test the impact of near native language (languages which have similar origin) samples and more closely associated tasks as well.
    
    \item Several instances showed that even MLMs failed due to inter-language interferences. These issues could potentially be addressed by integrating multilingual Named Entity Recognition (NER) \cite{vitiugin2024unraveling}.
    
    \item The pretraining of VMLMs can be more language inclusive, i.e., it should contain better data representation for code-mixed setting \cite{zhang-etal-2023-multilingual}. 
\end{itemize}

\section*{Ethics statement:}
All the datasets used in this paper are either publicly available or gathered directly from corresponding author with a permission to use for research purpose. No new data collection or annotation was done as part of our work, and hence we aren't releasing any dataset. It is important to note that the paper may contain offensive, mockery or discriminatory language towards certain individuals or groups. We acknowledge this and want to clarify that we do not agree with or support these views in any way. We strictly adhere to the Google Translate API's Terms of Service\footnote{\url{https://developers.google.com/terms}} for generating the translated Hindi datasets.


\bibliography{mybib}

\appendix

\section{Related works:}
\label{section:related_works}
In this section, we discussed the past works of this domain. We listed task-wise description of the past literature in the following subsections.


\subsection{Monolingual humor:}
The automatic detection of humor has gathered significant interest in NLP community. Most of the research in past literature focused on monolingual setting. From the task point of view, we saw binary classification (`humor' or `non-humor') \cite{mihalcea-strapparava-2005-making, purandare-litman-2006-humor, yang-etal-2015-humor, ramakrishna18_interspeech, hasan-etal-2019-ur, zhao-etal-2019-embedding, liu-hou-2023-mining}, multi-class and multi-label classification on humor targets \cite{HAHA-iberLEF-2021},  ranking (top 10 humorous utterances) \cite{zhao-etal-2019-embedding}, scoring (based on a reaction based humor score) \cite{yang-etal-2021-choral}, generation \cite{stock-strapparava-2005-hahacronym, Chen2024}, etc. From the approach point of view, past studies explored various approaches, ranging from statistical techniques to deep learning models \cite{abulaish2020survey}. For instance, initially researchers looked into stylistic features such as alliteration, antonyms, and adult slangs \cite{mihalcea-strapparava-2005-making}, and also prosody features like pitch, tempo, and emphasis \cite{purandare-litman-2006-humor}.  In addition, \cite{stock-strapparava-2005-hahacronym} used theoretical ideas like incongruity theory to generate humorous acronyms. Later, with the rise of deep-neural networks like RNNs, LSTMs, and CNNs with character ngram, Word2Vec and kNN features\cite{yang-etal-2015-humor, bertero-fung-2016-long, ramakrishna18_interspeech, hasan-etal-2019-ur}), research on computational humor identification has grown significantly. 
The development of attention mechanisms allowed identification context-based humor using transformer models \cite{weller-seppi-2019-humor} and pretrained language models \cite{yang-etal-2021-choral, shang-etal-2022-know, Chen2024}. 

\subsection{Code-mixed humor:}
Very few studies focused on humor detection in code-mixed environments, and most of them are present in the Indian context \cite{khandelwal-etal-2018-humor, sane-etal-2019-deep, 10.1145/3368567.3368576, 9442359}.
\citet{khandelwal-etal-2018-humor} introduced the first Hindi-English code-mixed dataset and evaluated it on statistical classifiers (like SVM, NB and Random Forest) with n-gram and bag-of-words features for humor detection in code-mixed setting. This dataset served as a valuable resource for subsequent studies in the field. \citet{sane-etal-2019-deep} proposed an attention-based bidirectional LSTM model using Continuous Bag of Words (CBOW) and Skip-gram embeddings for humor detection in the same code-mixed dataset.  Building upon existing research, \citet{10.1145/3368567.3368576} compared bidirectional LSTM and CNN models utilizing word and sentence embeddings for humor detection in Hindi-English code-mixed text. Their study provided insights into the comparative efficacy of different neural network architectures for this task. In a recent development, \citet{9442359} leveraged contextual attention mechanisms for multi-modal humor detection in Hindi-English code-mixed text. This approach highlighted the importance of considering contextual cues in code-mixed humor analysis.

\subsection{Monolingual sarcasm:}
Sarcasm detection, though a challenging task, has gained attention due to its repurcussions. It is also considered as an implicit hate.
From the task point of view, we saw many variations, like binary classification (`sarcastic' or `non-sarcastic') \cite{Tepperman-Joseph-Traum-2006, Tsur2010, joshi-etal-2015-harnessing, riloff-etal-2013-sarcasm}, sarcasm generation \cite{mishra-etal-2019-modular, zhao-etal-2023-multi, ilic-etal-2018-deep}, counter sarcasm generation \cite{peled-reichart-2017-sarcasm}, translation \cite{Sukmaningrum-2018}, etc. From an architectural point of view, researchers explored rule-based features such as prosodic, spectral, and contextual cues, along with patterns and punctuations \cite{Tepperman-Joseph-Traum-2006}, pattern matching by extracting High-Frequency Words (HFWs) and Content Words (CWs) \cite{Tsur2010} and shift of sentiment and various incongruities \cite{joshi-etal-2015-harnessing}. Later on, deep learning-based approaches gained pace and demonstrated promising results. The RNN, LSTM and CNN (or a combination of them) networks claimed to show strengths of semantic modelling \cite{ghosh-veale-2016-fracking, zhang-etal-2016-tweet}. Further, the combination of contextual pretrained embeddings and socio-linguistic features such as Named Entity Recognition (NER), Part-of-Speech (POS) tagging, Empath, and LIWC (Linguistic Inquiry and Word Count) features proved to be better at sarcasm classification \cite{patro-etal-2019-deep}. Later, the attention mechanism based pretrained transformers came into play to capture better contextual sarcastic cues from the sequence of text \cite{Potamias2020, babanejad-etal-2020-affective}.

\subsection{Code-mixed sarcasm:} Lately, the focus on code-mixed sarcasm detection started to come up \cite{10.1145/3632754.3633077, 9442359, aggarwal-etal-2020-really, shah-maurya-2021-effective}. Still the studies are limited in terms of datasets and architecture both. 
From the dataset point of view, we saw code-mixed sarcasm in Tamil-English and Malayalam-English \cite{10.1145/3632754.3633077}, Hindi-English \cite{swami2018corpus, aggarwal-etal-2020-really, DBLP:conf/esws/VijayBSAS18}, multimodal Hindi-English \cite{9442359}, etc. However, not all of the datasets are publicly available.
From the methodological point, \citet{aggarwal-etal-2020-really} uncovered a thorough comparison of deep learning based models like CNNs, LSTMs and Bi-directional LSTMs.
Finally, the advent of transformers allowed researchers to utilize sub-word level embeddings \cite{shah-maurya-2021-effective} and contextual attention mechanism for multi-modal Hindi-English code-mixed sarcasm detection \cite{9442359}.

\subsection{Multi-task learning:}
Multi-task learning has been used widely in the field of natural language processing for the past few years \cite{10.1145/3616855.3635690, 10.1145/3627704, tang-etal-2023-mvp}. From the task point of view, particularly in the code-mixed scenario, it has shown some promising results in various tasks such as stance detection \cite{sane-etal-2019-stance}, sentiment analysis \cite{wu-etal-2020-meistermorxrc}, cyberbullying detection \cite{10.1007/978-3-030-92273-3_36}, and emotion classification \cite{ameer-etal-2023-findings}. Architecture-wise, researchers utilized multi-channel CNN \cite{sane-etal-2019-stance}, pretrained BERT model \cite{wu-etal-2020-meistermorxrc}, BERT+VecMap \cite{10.1007/978-3-030-92273-3_36}, pretrained LMs and prompt tuning \cite{ameer-etal-2023-findings}.
However, the quantity of data has minimal impact on these tasks; instead, utilizing data from diverse sources has proven to be a more effective solution \cite{baranov-etal-2023-told}. Finetuning MLMs with task specific modules (e.g. adapters) achieved success in cross-lingual learning for low-resource languages \cite{pfeiffer-etal-2020-mad, parovic-etal-2022-bad}. 

\subsection{In-context learning:}
Although the very large language models are known for their knowledge and ability to perform well with just a few examples, their comprehension of low resource languages is still suboptimal \cite{bang-etal-2023-multitask}.
From the prompting point of view, \citet{liu-etal-2023-prompt} proposed prompt tuning, where they defined several prompt templates and verbalizers to assess whether the intended meaning of a comment contradicts the content provided in the prompt. Past work \cite{nag-etal-2024-cost} have compared the performance of native script and translated/ transliterated version of it for various tasks like sentiment classification, paraphrasing, intent classification, summarization, question answering, multichoice question answering, etc.

\subsection{Research gap: }
\citet{laureano-de-leon-etal-2024-code} reported that pretrained MLMs retained enough native language information for processing code-mixed text containing closely associated languages like Spanish and English. Thus, the MLMs improved in sentiment related tasks like code-mixed hate detection via native sample mixing \cite{mazumder2024improving}, but require native samples from the participating languages. \citet{choudhury-etal-2017-curriculum} also showed that a generic DNN performs best in LID and language modeling task when native samples are either mixed with code-mixed data or trained sequentially- first with native samples, followed by code-mixed data. 
As a result, past studies highlighted the benefits of native sample mixing and multi-task learning in various tasks. To the best of our knowledge, there is no existing study which uses both native samples and multitasking for code-mixed humor and sarcasm detection.

\subsection{Code-mixing:}

Code-mixing is a linguistic phenomenon common across multilingual speakers. Multilingual speakers are believed to outnumber monolingual speakers globally \cite{tucker1999global}. A large chunck of population in Asia, Europe, and North America know more then one language, this allows them to communicate among themselves by switching languages in a single utterance. It is more prominent in informal communications like social media posts and voice mails \cite{patro2017all}. 

\section{Additional dataset details:}
\label{dataset_details}


\begin{figure*}
    \centering
        \includegraphics[trim=2.5cm 5.7cm 2.5cm 2.5cm, clip, width=\textwidth]{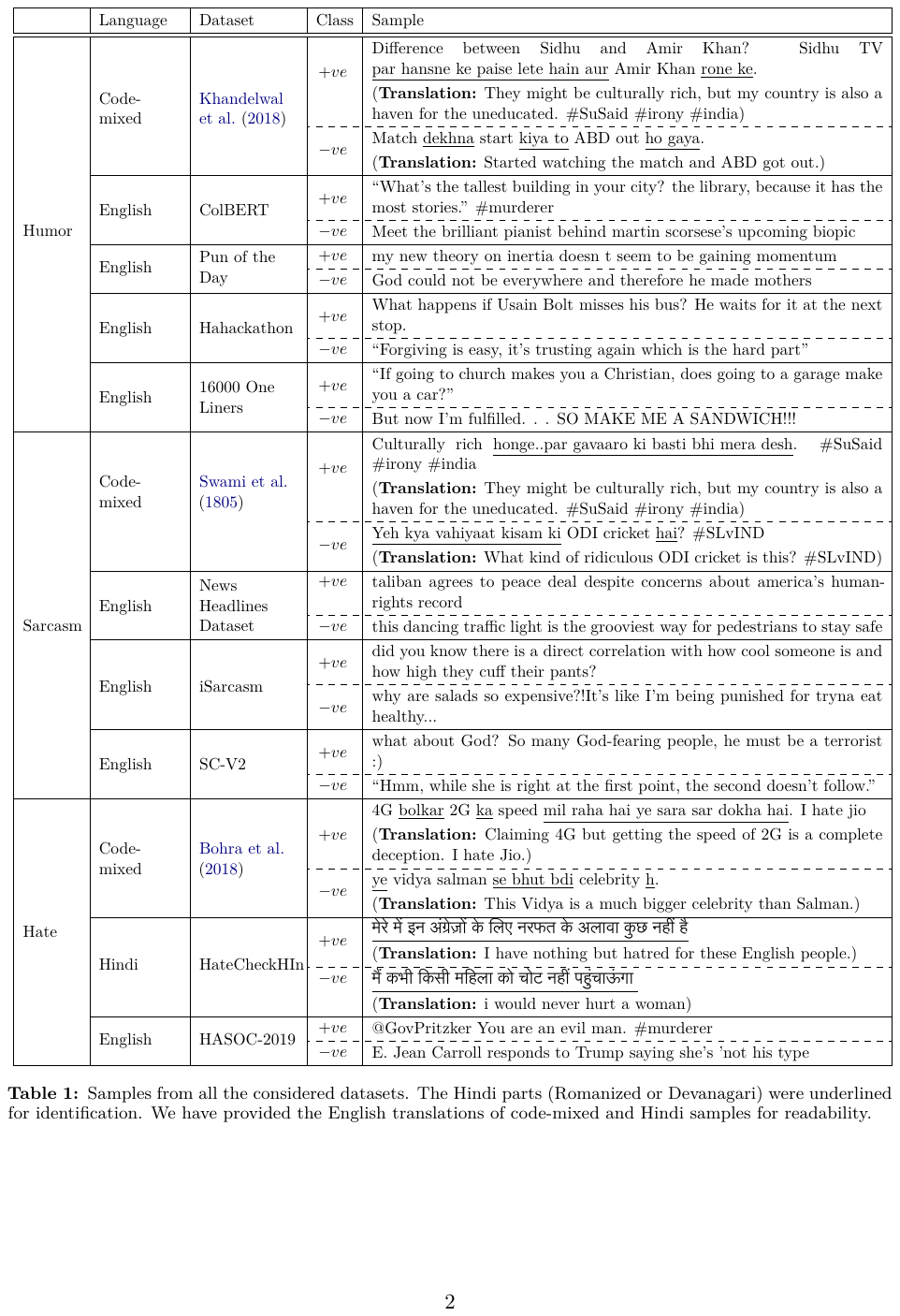}
    \caption{Samples from all the datasets. The Hindi parts (Romanized or Devanagari) were underlined for identification. We have provided the English translations of code-mixed and Hindi samples for readability.}
    \label{fig:dataset_samples_table}
\end{figure*}

In this section, we described the details of code-mixed and monolingual datasets considered in our study. They were arranged task-wise in the following subsections.  

\subsection{Humour:} We could find only one publicly available Hindi-English code-mixed humour dataset, introduced by \textbf{\citet{khandelwal-etal-2018-humor}} (GPL 3.0 licensed). However, there are several English humour datasets proposed in the literature \cite{annamoradnejad2024colbert, yang-etal-2015-humor, meaney-etal-2021-semeval,mihalcea-strapparava-2005-making, weller-seppi-2020-rjokes, tang2022naughtyformertransformerunderstandsoffensive, kamal2020self}. Out of which, we considered ColBERT (\textbf{Col}) \cite{annamoradnejad2024colbert}, Pun of the Day (\textbf{POTD}) \cite{yang-etal-2015-humor}, HaHackathon (\textbf{HaHa}) \cite{meaney-etal-2021-semeval} and 16000 One Liners (\textbf{16000}) \cite{mihalcea-strapparava-2005-making} datasets. This is because they are relatively balanced and widely used in past studies\cite{kenneth2024systematic}. On the contrary, we couldn't find any appropriate Hindi humour dataset. Although \citet{kumar2023humourhindinet} recently proposed one, it comprised dialogues taken from a famous Hindi TV series. Further, upon deep inspection, we found that the associated labels depend highly on preceding dialogues with varying contexts (dialogue iterations). Thus, we restricted ourselves from using them. In the following, we report a brief description of the individual datasets,
\subsubsection{Code-mixed dataset: } 
\begin{itemize}
    \item \textbf{\citet{khandelwal-etal-2018-humor}:} \citet{khandelwal-etal-2018-humor} gathered 10,478 tweets from various domains like `sports', `entertainment', `politics', etc. They manually identified 3,453 code-mixed tweets by discarding monolingual Hindi and English tweets. Each identified tweet was annotated by three language experts, skilled in both Hindi and English. They tagged each tweet as ``humorous'' (H) or ``non-humorous'' (N). As a norm, they labelled the tweets containing anecdotes, fantasy, irony, jokes, and insults as humorous, and the tweets with facts, dialogues and speeches that didn't provoke any laughter were labelled non-humorous. We found that, over time, some of the tweets in the original dataset got deleted. Therefore, we resorted to using the currently available samples, which is a total of 2,951 tweets (1,759 humorous and 1,192 non-humorous).
\end{itemize}


\subsubsection{Native language datasets:}
\begin{itemize}
    \item \textbf{ColBERT (Col)\cite{annamoradnejad2024colbert}:} ColBERT \cite{annamoradnejad2024colbert} was formed by combining samples from two previously published datasets: (i) news website \cite{misra2022news} and (ii) jokes website \cite{weller-seppi-2019-humor}. The news website dataset consists of 200k Huffington Post news headlines from 2012-2018. They are from different categories like politics, wellness, entertainment and parenting. The jokes website dataset consists of around 231k humorous samples collected from two subreddits : /r/jokes and /r/cleanjokes. Their authors randomly selected 100k samples from both datasets after a few fine-grained preprocessing steps including de-duplication of samples, lexical statistics matching and title case formatting.
    \item \textbf{Pun of the Day (POTD)\cite{yang-etal-2015-humor}:}  The Pun of the Day \cite{yang-etal-2015-humor} (MIT licensed) dataset consists of humorous samples directly collected from the pun-of-the-day jokes website\footnote{\url{http://www.punoftheday.com}}. The non-humorous samples were scraped from AP news, the New York Times, Yahoo! Answer and Proverb websites. Their authors performed a curated sampling of negative samples to minimize domain differences. 

    \item \textbf{HaHackathon (HaHa)\cite{meaney-etal-2021-semeval}:} The HaHackathon \cite{meaney-etal-2021-semeval} comprised of 10k samples. \citet{meaney-etal-2021-semeval} created this dataset with Twitter posts (80\%) and Kaggle Short Jokes samples\footnote{\url{https://www.kaggle.com/abhinavmoudgil95/short-jokes}} (20\%).
    While the humorous tweets were collected from humorous Twitter accounts (e.g. @humurous1liners and @conanobrien), and the non-humorous tweets were collected from some celebrity accounts (e.g. @thatonequeen and @Oprah). 
    From the Kaggle dataset, they selected samples expressing humour and offence.
    The accumulated 10k samples were annotated by twenty US-based annotators aged between 18 and 70 years, by answering the question \textit{``Is the intention of this text to be humorous?''}.
    \item \textbf{16000 One Liners (16000)\cite{mihalcea-strapparava-2005-making}:} This dataset \cite{mihalcea-strapparava-2005-making} consists of 32k short sentences. Out of which, 16k are humorous samples. They were automatically collected through a web-based bootstrapping process. The remaining 16k samples are non-humorous, and they were collected from Reuters titles, Proverbs and British National Corpus (BNC).
\end{itemize}

\subsection{Sarcasm:} We could find two publicly available Hindi-English code-mixed sarcasm datasets. They were introduced by \textbf{\citet{swami2018corpus}} and \citet{shah2022effective}. Out of them, \citet{shah2022effective} distantly labelled their samples with the help of hashtags associated with tweets.
We restricted ourselves from using it as our primary objective is to improve code-mixed sarcasm and introducing this dataset can make the results noisy. 
There are several native English sarcasm datasets present as well \cite{misra2019sarcasm, joshi-etal-2016-challenging, abu-farha-etal-2022-semeval, ptacek-etal-2014-sarcasm, oprea-magdy-2020-isarcasm, lukin-walker-2013-really, oraby-etal-2016-creating}. In the present work, we considered to experiment with News Headlines Dataset (\textbf{NHD}) \cite{misra2019sarcasm}, iSarcasm (\textbf{iSarc}) \cite{abu-farha-etal-2022-semeval} and Sarcasm Corpus V2 (\textbf{SC-V2} hence after) \cite{oraby-etal-2016-creating}; as they were widely studied in literature\cite{joshi2017automatic, chen2024survey}. iSarcasm and SC-V2 were manually annotated by expert annotators, whereas the News Headlines Dataset is a distant labelled dataset. Similar to the case of Hindi humour detection, we could not find any publicly available Hindi sarcasm dataset. In the following, we provided a brief description of the considered sarcasm datasets:
\subsubsection{Code-mixed dataset: } 
\begin{itemize}
    \item \textbf{\citet{swami2018corpus}:} To create this dataset (GPL-3.0 licensed), \citet{swami2018corpus} scrapped tweets containing keywords `politics', `cricket', and `Bollywood'. They manually filtered Hindi-English code-mixed tweets by inspecting individual samples. Tweets containing   `\#sarcasm' and `\#irony' and lacking them are kept in the initial pool of sarcastic and non-sarcastic samples, respectively. A group of language experts well-versed in Hindi and English annotated the samples with an inter-annotator agreement \cite{fleiss1973equivalence} of 0.79. The final version of their dataset has 5,250 samples, out of which 504 are sarcastic. 
    

    
\end{itemize}

\subsubsection{Native language datasets:}
\begin{itemize}
    \item \textbf{News Headlines Dataset (NHD)\cite{misra2019sarcasm}:} The samples collected in this dataset were headlines from two websites: (i) TheOnion\footnote{\url{https://www.theonion.com/}} briefs, comprises of sarcastic explanations of current events (as sarcastic samples) and (ii) HuffPost\footnote{\url{https://www.huffpost.com/}}, an American news website (as non-sarcastic samples). They down-sampled HuffPost samples to nearly match the sarcastic samples, resulting in a balanced dataset of nearly 26.7k samples.
    
    \item \textbf{iSarcasm (iSarc)\cite{abu-farha-etal-2022-semeval}:} This dataset consists of 5,735 tweets implicitly labelled by tweet authors. \citet{abu-farha-etal-2022-semeval} conducted a survey among  English speakers having Twitter accounts. Participants were asked to provide one link to their sarcastic and three links to their non-sarcastic tweets posted in their recent past. Additionally, the authors also requested the survey participants to provide a non-sarcastic version of their sarcastic tweets.  
    \item \textbf{SC-V2\cite{oraby-etal-2016-creating}:} This dataset consists of 9,386 text samples collected from three different online debate forums like \url{4forums.com}, \url{CreateDebate.com} and \url{Convinceme.Net}. Nine expert annotators then annotated each sample as `sarcastic' or `not-sarcastic'. Further, annotators also labelled them for three sub-types of sarcasm: general, hyperbole and rhetorical questions. This dataset is a subset of the Internet Argument Corpus (IAC) \cite{walker-etal-2012-corpus}.
\end{itemize}

\subsection{Hate:} To facilitate knowledge sharing across tasks in MTL frameworks, we used publicly available Hindi-English code-mixed and native (i.e. monolingual Hindi and English) hate datasets. We could find only one Hindi \citet{das2022hatecheckhin} (hereafter referred to as \textbf{HCHIn}) and Hindi-English code-mixed  \textbf{\citet{bohra-etal-2018-dataset}} hate dataset that is publicly available. Further, as an English hate dataset, we used HASOC-2019 (English) (\textbf{HASOC} hence after)\footnote{HASOC-2019: \url{https://hasocfire.github.io/hasoc/2019/dataset.html}\label{refnote}} as it is widely used in past works. In the following, we provide a brief description of the individual datasets:
\subsubsection{Code-mixed dataset: } 
\begin{itemize}
    \item \textbf{\citet{bohra-etal-2018-dataset}:} To create this dataset, authors retrieved 112,718 tweets based on a predefined list of hashtags and keywords related to `politics', `public protests', `riots', etc. Following this, 4,575 code-mixed tweets were manually filtered, and two expert annotators tagged them as ``hate''(H) or ``non-hate''(NH).
\end{itemize}
\subsubsection{Native language datasets:}
\begin{itemize}
    \item \textbf{HateCheckHIn (HCIn)\cite{das2022hatecheckhin}:}  This dataset contains 4,754 Hindi samples, each annotated with `hate' or `non-hate' by expert annotators well-versed in the Hindi language. This dataset was constructed to test the weaknesses of Hindi hate speech detection models. \citet{das2022hatecheckhin} manually designed 28 monolingual functionality tests for that purpose. The quality of the test cases was verified by two expert annotators. 
    \item \textbf{HASOC-2019\footref{refnote} (HASOC):} It contains 5,852 social media posts collected from Twitter and Facebook using hashtags and keywords. Following this, each sample was annotated as `hate' or `non-hate' by organizers of the HASOC track.
\end{itemize}


\section{Additional baseline details:}
\label{addn_baseline_details}
In this section, we described the details of baseline methods considered in our study. They were arranged task-wise in the following subsections.

\subsection{Humor:}
\label{addn_baseline_details_humour}
\begin{itemize}
    \item \citet{agarwal-narula-2021-humor-generation} experimented with various neural architectures ranging from variants of LSTMs (such as vanilla LSTM, Bi-LSTM and Bi-LSTM with attention mechanism) to MLMs like mBERT \cite{devlin-etal-2019-bert} and IndicBERT \cite{kakwani-etal-2020-indicnlpsuite}. They found that MLMs by far outperform the LSTMs in terms of accuracy. IndicBERT is pre-trained on 12 major Indian languages (which includes 1.84 B Hindi tokens) with fewer parameters than mBERT \cite{devlin-etal-2019-bert}. We considered both language models as baseline.

    \item \citet{muttaraju-etal-2022-semi-supervised} approached the problem in a semi-supervised manner. They used a ratio of 1:100 for labeled versus unlabeled data. The labeled subset was used to train a classifier at first, and then they utilized the same classifier to get pseudo-labels from the unlabeled data points based on a threshold of prediction probability. The new training set for supervised modeling now consists of both pseudo-labeled and gold-labeled samples. This process is repeated until either the maximum number of iterations is reached or no more labeled data remains. From modeling point of view, they utilized HinglishBERT\footnote{\url{https://huggingface.co/nirantk/hinglish-bert}} within the GAN-BERT\cite{croce-etal-2020-gan} architecture.
\end{itemize}

 \subsection{Sarcasm:}
 \label{addn_baseline_details_sarcasm}
\begin{itemize}
    \item \citet{pandey2023bert} experimented on a variety of neural architectures ranging from linear layers, CNNs, and LSTMs to pre-trained BERT\cite{devlin-etal-2019-bert} and BERT-LSTM (LSTM stacked upon mBERT). LSTM-BERT significantly outperformed the others in terms of F1 score of positive class. Thus, we considered it as our baseline.
    \item \citet{aloria2023hilarious} preprocessed the individual samples using a spelling-checker\footnote{\url{https://pypi.org/project/pyspellchecker/}}. They experimented with a variety of architectures like CNNs, bi-LSTMs, statistical ensemble classifiers, BERT-LSTM and a novel BERT-GRU (bi-directional GRU stacked upon mBERT) architecture. They found that BERT-GRU significantly outperforms others in terms of the F1 score of the positive class.
\end{itemize}

\section{Additional details of Exp. 3:}

\subsection{Prompting details:}
\label{section:prompts}

\begin{figure}[H]
    \centering
    \includegraphics[trim=0cm 4.2cm 0cm 0cm, clip, width = \columnwidth]{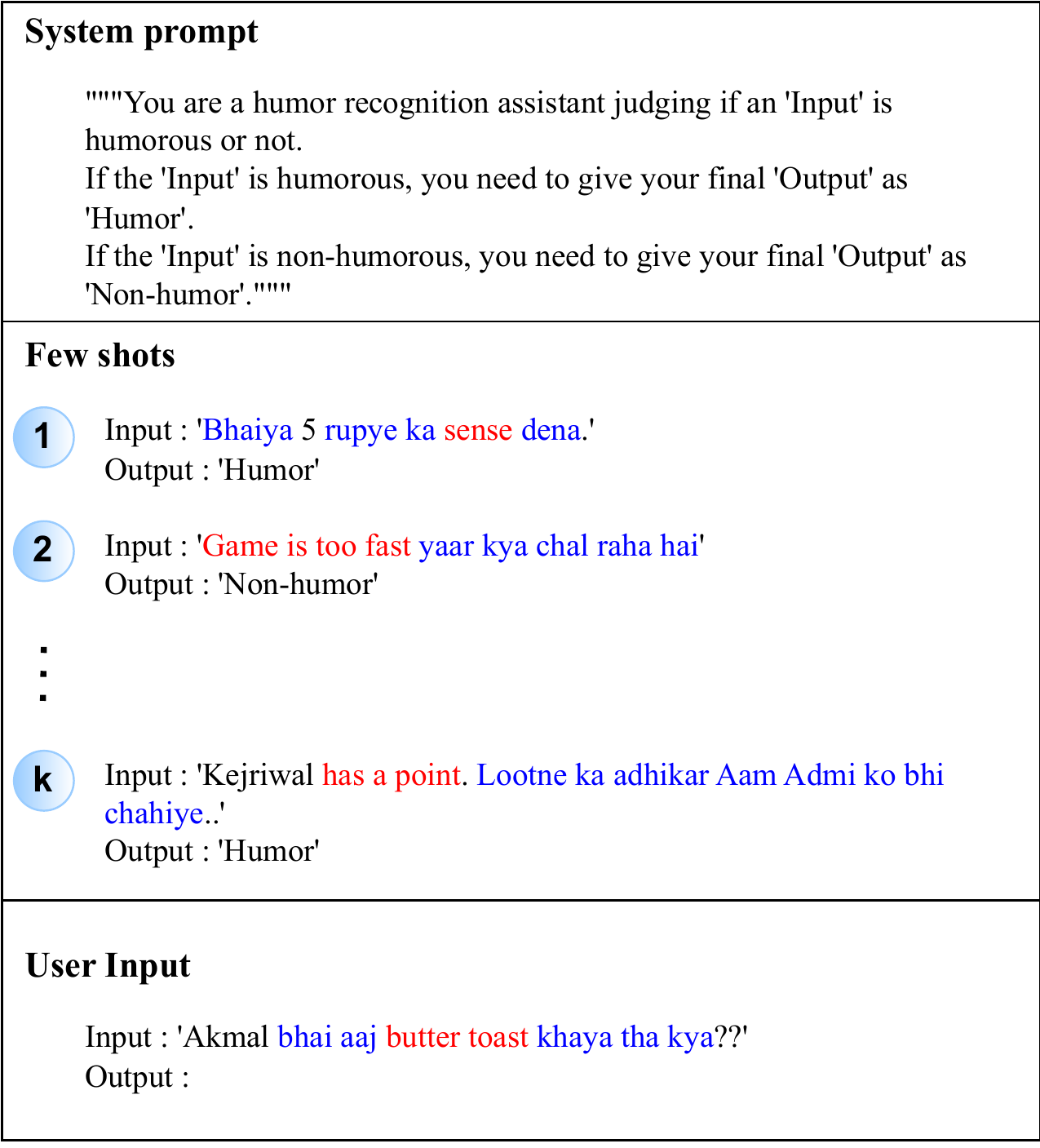}
    \caption{ Prompt template for k-shot prompting utilized for humor detection. English parts are marked in red and the Hindi parts are marked in blue.}
    \label{fig:prompt}
\end{figure}

In this section, we described the prompting details for in-context learning for conducting \textbf{Exp. 3} (refer Section \ref{subsection:VMLM}). Our prompt consisted of three parts, i) system prompt, where we explained the task, ii) few shots, examples that we fed to learn from, and iii) user input, which is the query for which we needed the predicted label.
We selected the few shots through clustering technique which is considered to be a better approach \cite{huzaifah-etal-2024-evaluating} than randomly picking examples. We presented the prompt template in Figure \ref{fig:prompt}. The detailed results for the first scenario of few-shot prompting VMLM (Experiment-3) is presented in Table-\ref{tab:results_vmlm_all_fewshot}. The VMLMs were prompted with 0-shot and k-shot examples given in the context. The user input query remained in code-mixed language in each of the cases.

\afterpage{
\begingroup
\renewcommand{\arraystretch}{1.08} 
\begin{table}
    \centering
    \setkeys{Gin}{keepaspectratio}
    \resizebox*{\columnwidth}{\textwidth} {
    \begin{tabular}{ll|ccccc}
    \Xhline{3\arrayrulewidth}
        \multicolumn{7}{c}{\textbf{\Large{Humor}}}  \\
        \Xhline{3\arrayrulewidth}
        Model & Dataset & 0-shot & 2-shot & 4-shot & 8-shot & 12-shot \\ \hline \hline
        Gemma & CM & \textbf{0.09} & 0.08 & 0.08 & \textbf{0.09} & 0.07 \\
         & Col &  & 0.08 & \textbf{0.09} & 0.04 & 0.04 \\
         & POTD &  & 0.04 & 0.03 & \textbf{0.07} & \textbf{0.07} \\
         & HaHa &  & 0.10 & \textbf{0.11} & 0.02 & 0.10 \\
         & 16000 &  & \textbf{0.10} & 0.02 & 0.04 & 0.02 \\ \hline
        Aya Expanse & CM & 0.72 & 0.72 & 0.72 & \textbf{0.73} & 0.72 \\
         & Col &  & \textbf{0.74} & 0.71 & 0.73 & 0.69 \\
         & POTD &  & 0.70 & \textbf{0.73} & 0.72 & 0.68 \\
         & HaHa &  & \textbf{0.71} & 0.68 & 0.67 & 0.63 \\
         & 16000 &  & 0.72 & \textbf{0.73} & \textbf{0.73} & 0.72 \\ \hline
        Llama-3.1 & CM & \textbf{0.75} & \textbf{0.75} & 0.74 & \textbf{0.75} & \textbf{0.75} \\
         & Col &  & 0.72 & \textbf{0.74} & 0.68 & 0.63 \\
         & POTD &  & 0.75 & \textbf{0.76} & 0.73 & 0.65 \\
         & HaHa &  & \textbf{0.75} & \textbf{0.75} & 0.55 & 0.62 \\
         & 16000 &  & 0.67 & \textbf{0.77} & 0.64 & 0.63 \\ \hline
        GPT-4 & CM & 0.73 & 0.73 & \textbf{0.74} & \textbf{0.74} & 0.73 \\
         & Col &  & 0.47 & 0.51 & 0.56 & \textbf{0.57} \\
         & POTD &  & 0.50 & 0.55 & \textbf{0.62} & 0.56 \\
         & HaHa &  & 0.55 & 0.50 & \textbf{0.56} & 0.55 \\
         & 16000 &  & 0.40 & 0.38 & 0.47 & \textbf{0.58} \\ \hline

    \end{tabular}
    }

    \vspace{3mm}

    \setkeys{Gin}{keepaspectratio}
    \resizebox*{\columnwidth}{\textwidth} {
    \begin{tabular}{ll|ccccc}
    \Xhline{3\arrayrulewidth}
        \multicolumn{7}{c}{\textbf{\Large{Sarcasm}}}  \\
        \Xhline{3\arrayrulewidth}
        Model & Dataset & 0-shot & 2-shot & 4-shot & 8-shot & 12-shot \\ \hline \hline
        Gemma & CM & 0.30 & 0.29 & 0.30 & \textbf{0.34} & 0.21 \\
         & NHD &  & \textbf{0.11} & \textbf{0.11} & \textbf{0.11} & 0.02 \\
         & iSarc &  & 0.35 & \textbf{0.39} & 0.37 & 0.30 \\
         & SC-V2 &  & 0.18 & \textbf{0.47} & \textbf{0.47} & 0.18 \\ \hline
        Aya Expanse & CM & 0.18 & \textbf{0.21} & 0.20 & \textbf{0.21} & 0.18 \\
         & NHD &  & \textbf{0.21} & 0.20 & \textbf{0.21} & 0.20 \\
         & iSarc &  & 0.21 & 0.21 & \textbf{0.26} & 0.21 \\
         & SC-V2 &  & 0.21 & 0.21 & \textbf{0.23} & 0.21 \\ \hline
        Llama-3.1 & CM & 0.18 & 0.18 & 0.18 & \textbf{0.21} & 0.18 \\
         & NHD &  & 0.30 & \textbf{0.45} & 0.37 & 0.36 \\
         & iSarc &  & 0.27 & 0.31 & 0.32 & \textbf{0.51} \\
         & SC-V2 &  & 0.24 & 0.26 & 0.22 & \textbf{0.28} \\ \hline
        GPT-4 & CM & 0.76 & 0.77 & 0.76 & \textbf{0.78} & 0.75 \\
         & NHD &  & 0.12 & 0.16 & 0.16 & \textbf{0.17} \\
         & iSarc &  & 0.20 & 0.23 & \textbf{0.25} & 0.24 \\
         & SC-V2 &  & 0.40 & \textbf{0.43} & 0.42 & 0.41 \\ \hline

    \end{tabular}
    }

    \caption{Results of prompting VMLMs with in-context few shot examples. Here, 0-shot represents providing no example in the context of the prompt, and k-shot represents k examples of input-output pairs are given in the context of the prompt. Notation: CM for code-mixed.}
    \label{tab:results_vmlm_all_fewshot}
\end{table}
\endgroup
}

\subsection{LoRA-adapter based finetuning:}
For the second scenario, we performed instruction finetuning on the VMLMs using the default parameters of LoRA adapter based supervised finetuning given in LLaMA-Factory \cite{zheng2024llamafactory}. We utilized the same prompt template for giving instruction to the models.

\subsection{Detailed observations from Exp. 3:}
\label{Detailed Exp. 3 Results}

In this section, we reported the additional observations of third experiment. While investigating the poor performance of the VMLMs, we identified certain patterns. The VMLMs were prompted to provide reasoning for their label predictions. Here are some observations,

    \begin{itemize}
        \item VMLMs favored specific labels like humour and sarcasm, where it extracted comedic or ironic effect in plain non-humourous statements. For instance, in the non-humourous sample \textit{`\textcolor{blue}{So jao sab}, \textcolor{blue}{kal} \textcolor{red}{Monday} \textcolor{blue}{hai}.'} (\textbf{Gloss:} `Go to sleep, everyone. Tomorrow is Monday.'), Llama-3.1 predicted it as humourous and stated the reason : \textit{`The input is a Hindi phrase that translates to ``So go everyone, it's Monday." The humour lies in the fact that it's a common expression that people use to bid farewell on Fridays, but it's being used on Monday, which is the opposite day. It's a play on expectations and a lighthearted way to poke fun at the monotony of the workweek.'}.
        
        \item We could find instances where VMLM assumed some prior context over the original non-sarcastic sample \textit{`\textcolor{blue}{Kabhi} \textcolor{red}{triple} \textcolor{blue}{talaq halala pe v kuch} \textcolor{red}{post} \textcolor{blue}{kardo agar himmat hai to}'} (\textbf{Gloss:} `Try posting something about triple talaq and halala if you have the courage.'), GPT-4 predicted it as sarcastic and stated the reason as \textit{`The statement is sarcastic as it challenges the recipient’s courage (``agar himmat hai to"), implying reluctance or hypocrisy. It uses a mocking tone to suggest selective bias in posting about issues. The indirect criticism and taunting nature make it a clear case of sarcasm.'}
        
        \item VMLMs showed a tendency to pick keywords, like in the non-sarcastic example \textit{`\textcolor{blue}{Yeh bhee} \textcolor{red}{cricket} \textcolor{blue}{kee deewani hain}....'} (\textbf{Gloss:} `She is also crazy about cricket...'), Llama-3.1 predicted it sarcastic and gave the reason as \textit{`The phrase ``Yeh bhee cricket kee deewani hain" translates to "You are also a cricket fanatic" in English. However, the use of the word \textbf{``bhee"} (also) in a sentence suggests that the speaker is being sarcastic, implying that the person is extremely obsessed with cricket, which is often considered a stereotypical trait in Indian culture. The tone of the sentence suggests that the speaker is mocking or ridiculing the person's enthusiasm for cricket.'}
    \end{itemize}


\section{Experimental configurations: }
\label{config}

In this section, we reported the details of experimental setups and model configurations. 

\subsection{Experimental set-up:}
\label{experimental_config}

\begin{figure*}[!htbp]
    \centering
    \includegraphics[width=1.8\columnwidth]{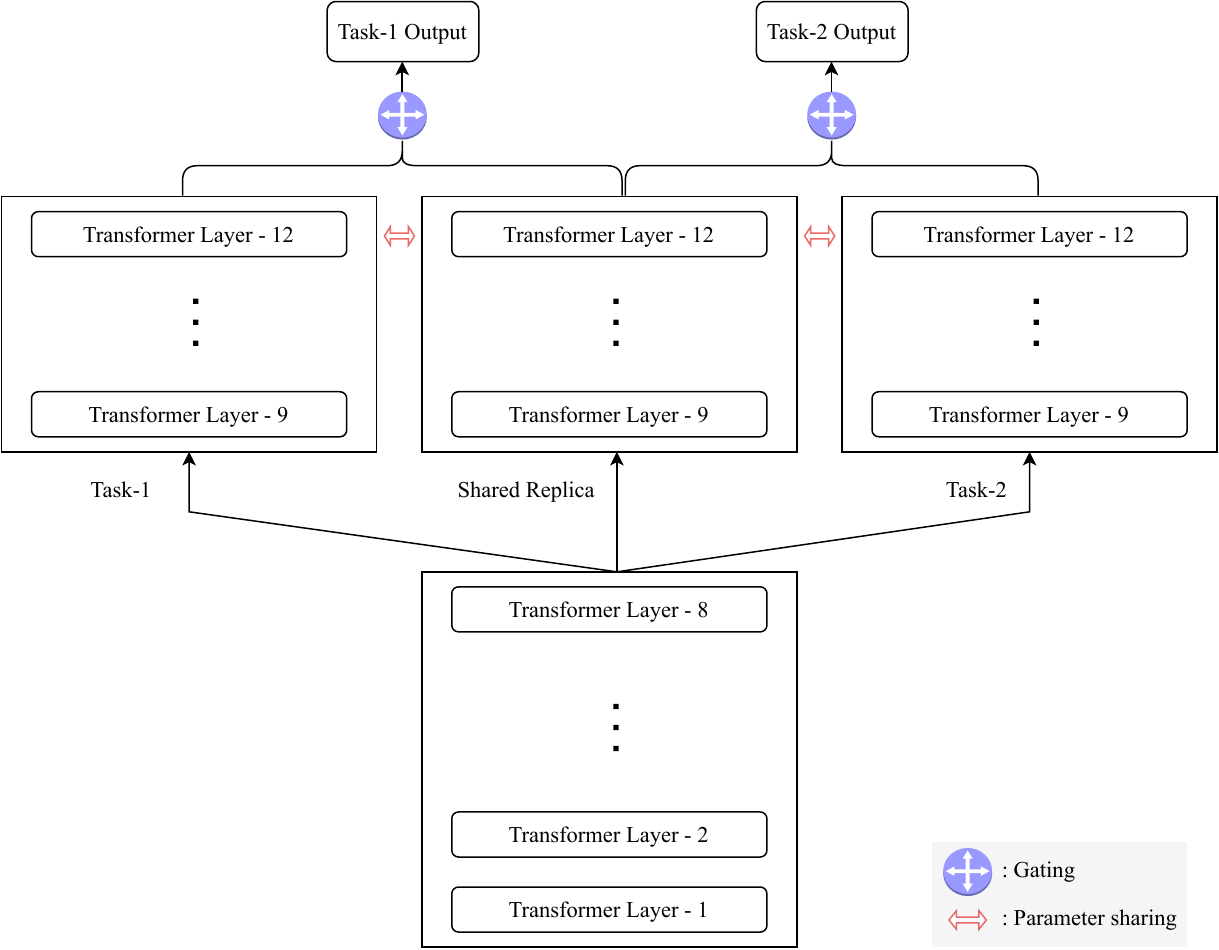}
    \caption{ MTL architecture when number of tasks is two, i.e., $T=2$.}
    \label{fig:MTL_arch}
\end{figure*}

\begin{figure}[H]
    \centering
    \includegraphics[width=\columnwidth]{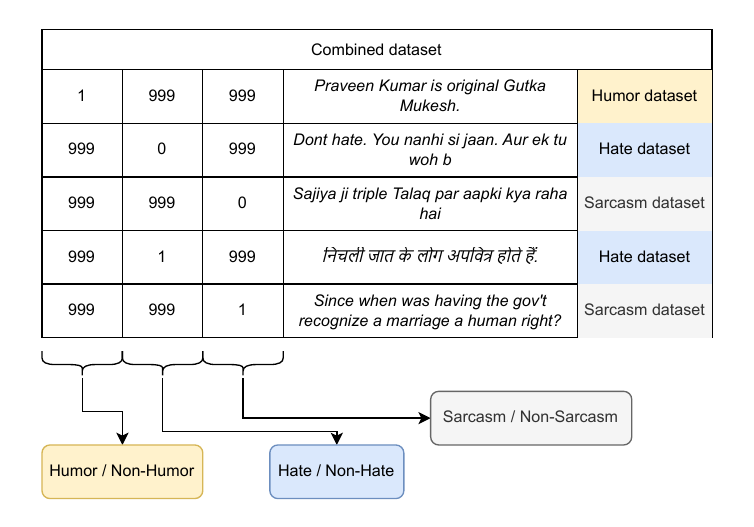}
    \caption{ Restructured dataset considered for MTL-based experiments.}
    \label{fig:dataframe}
\end{figure}

To conduct our experiments, we divided the code-mixed datasets into a training(80\%), validation(10\%) and test(10\%) set, with stratified sampling. We kept the test set the same as in \cite{khandelwal-etal-2018-humor} and \cite{swami2018corpus}. Additionally, we constructed several augmented training sets comprising native language task samples. Since no Hindi humour and sarcasm datasets were readily available, we created synthetic datasets by translating some portions of English datasets using Google Translator API. We randomly sampled data points from English and candidate translation samples.
Further, we ensured an equal number of samples to be selected from both classes in each augmented training set to avoid complexities related to class imbalance during training. Table \ref{tab:dataset_stat_used} reports the label distribution of the considered training sets, validation sets and test sets. For our Multi-Task Learning (MTL) experiment, we trained our model by feeding samples batchwise, with each batch containing samples from multiple tasks. This approach avoided sequential training to prevent bias towards any specific task, especially the one processed last. As a result, samples from humor, sarcasm and hate detection appeared in the same batch. For cases where the task label was missing, we used an ignore label (`$999$'). Figure \ref{fig:dataframe} provides an example dataframe. The main architecture of our MTL framework is presented in Figure \ref{fig:MTL_arch}. This is a BERT-based architecture consisting of 12-layers. The whole model is divided into two halves: i) bottom 8 layers common for all tasks and ii) top 4 layers for task-specific training. Here, the upper module consisting of top 4 layers is thus replicated n times, where $n$ is the number of tasks added with one. This extra upper module is for shared features among the semantically related tasks. Finally, for each task, a gating mechanism combines the shared module output with the task-specific module output, to get the final logits. For clarity in each step, we also provided the related pseudocode in Algorithm \ref{alg:shared-cross-task-model}.

\begingroup
\renewcommand{\arraystretch}{1.2} 
\begin{table}[ht]
    \centering
    \setkeys{Gin}{keepaspectratio}
    \resizebox*{\columnwidth}{1.0\textheight} {
    \begin{tabular}{|l|l|c|c|c|}
    \hline
        Partition & Dataset & \# Humor & \# Non-Humor \\ \hline \hline
        Train & Code-mixed & 1407 & 953 \\ \hline
        \multirow{4}{4em}{Augment English} & Col & 1180 & 1180 \\ \cline{2-4}
        ~ & POTD & 1180 & 1180 \\ \cline{2-4}
        ~ & HaHa & 1180 & 1180 \\ \cline{2-4}
        ~ & 16000 & 1180 & 1180 \\ \hline
        \multirow{4}{4em}{Augment Hindi} & Col (translated) & 1180 & 1180 \\ \cline{2-4}
        ~ & POTD (translated) & 1180 & 1180 \\ \cline{2-4}
        ~ & HaHa (translated) & 1180 & 1180 \\ \cline{2-4}
        ~ & 16000 (translated) & 1180 & 1180 \\ \hline
        Val & Code-mixed & 176 & 119 \\ \hline
        Test & Code-mixed & 176 & 119 \\ \hline
    \end{tabular}}

\vspace{10pt} 

    \setkeys{Gin}{keepaspectratio}
    \resizebox*{\columnwidth}{1.0\textheight} {
    \begin{tabular}{|l|l|c|c|}
    \hline
        Partition & Dataset & \# Sarcasm & \# Non-Sarcasm \\ \hline \hline
        Train & Code-mixed & 403 & 3797 \\ \hline
        \multirow{4}{4em}{Augment English} & NHD & 2100 & 2100 \\ \cline{2-4}
        ~ & iSarc  & 1067 & 1067 \\ \cline{2-4}
        ~ & SC-V2 & 2100 & 2100 \\ \hline
        \multirow{4}{4em}{Augment Hindi} & NHD (translated) & 2100 & 2100 \\ \cline{2-4}
        ~ & iSarc (translated) & 1067 & 1067 \\ \cline{2-4}
        ~ & SC-V2 (translated) & 2100 & 2100 \\ \hline
        Val & Code-mixed & 50 & 475 \\ \hline
        Test & Code-mixed & 50 & 475 \\ \hline
    \end{tabular}}

\vspace{10pt} 

    \setkeys{Gin}{keepaspectratio}
    \resizebox*{\columnwidth}{1.0\textheight} {
    \begin{tabular}{|l|l|c|c|}
    \hline
        Partition & Dataset & \# Hate & \# Non-Hate \\ \hline \hline
        Train & Code-mixed & 1661 & 2914 \\ \hline
        \multirow{2}{4em}{Augment train} & HCIn & 1416 & 1416 \\ \cline{2-4}
        ~ & HASOC & 2261 & 2261 \\ \hline
    \end{tabular}}

\caption{Dataset statistics considered for the native sample mixing experiments with their train-val-test split. Notation: \# for number of samples,`translated' for translated Hindi.}
\label{tab:dataset_stat_used}
\end{table}
\endgroup

\begin{algorithm}[H]
  \caption{MultiTaskModel Algorithm}
  \label{alg:shared-cross-task-model}
  \textbf{Input:} Text input tokens ($input$) \\
  \textbf{Given:} BERT encoder ($BERT$), Taskwise last four layers of BERT module ($module_{task_{1}}, module_{task_{2}}$), Gating scheme ($gate$) \\
  \textbf{Output:} Logits ($comb_{task_{1}}, comb_{task_{2}}$).
  \begin{algorithmic}[1]
    \Function{MultiTaskModel}{$input$}
      \State $bert_{hidden} \gets BERT(input)$ 
      
      \State $bottom \gets bert_{hidden}[8]$
      
      \For{$layer$ in $module_{task_{1}}$}
        \State $task_{1} \gets layer(bottom)$
      \EndFor

      \For{$layer$ in $module_{task_{2}}$}
        \State $task_{2} \gets layer(bottom)$
      \EndFor

      \State $comb_{task_{1}} \gets gate(bert_{hidden}, task_{1})$
      \State $comb_{task_{2}} \gets gate(bert_{hidden}, task_{2})$
      
      \State \Return $comb_{task_{1}}, comb_{task_{2}}$
    \EndFunction
  \end{algorithmic}
\end{algorithm}



For each task $t \in \{1,2\}$, the final task-specific representation is computed by gating the shared BERT representation ($\mathbf{h}_{\text{BERT}}$) with the task-specific features ($\mathbf{h}_{\text{task}_t}$):

\begin{align}
\mathbf{o}_t &= \text{Gate}(\mathbf{h}_{\text{BERT}}, \mathbf{h}_{\text{task}_t}) \nonumber \\
&= \boldsymbol{\alpha}_t \odot \mathbf{h}_{\text{BERT}}
+ (1-\boldsymbol{\alpha}_t) \odot \mathbf{h}_{\text{task}_t}
\end{align}

where the gate coefficient $\boldsymbol{\alpha}_t$ is computed as:
\begin{align}
\boldsymbol{\alpha}_t &= \sigma\big(
\mathbf{W}_{g,t}[
\mathbf{h}_{\text{BERT}} \Vert 
\mathbf{h}_{\text{task}_t}] + 
\mathbf{b}_{g,t}\big)
\end{align}

Here:
\begin{itemize}
    \item $\mathbf{h}_{\text{BERT}}$: Final hidden states from shared replica of BERT.
    \item $\mathbf{h}_{\text{task}_t}$: Task-specific hidden representation after processing through $\text{module}_{task_t}$.
    \item $\sigma$: Sigmoid activation $\in (0,1)$.
    \item $\mathbf{W}_{g,t} \in \mathbb{R}^{D \times 2D}, \mathbf{b}_{g,t} \in \mathbb{R}^{D}$: Gating parameters for task $t$.
\end{itemize}

To implement the soft-parameter sharing, we introduced a regularization term in the joint loss function. The joint loss function for the MultiTaskModel (when number of tasks, i.e., $T=2$) is defined as:
\[
\mathcal{L}_{\text{joint}} = L_1 + L_2 + \lambda \cdot \|\mathbf{W}_{task_{1}} - \mathbf{W}_{task_{2}}\|_2
\]

where:
\begin{itemize}
    \item \( L_1 \): Loss for Task 1,
    \item \( L_2 \): Loss for Task 2,
    \item \( \lambda \): Regularization strength,
    \item \( \mathbf{W}_{task_{1}} \): Weight matrix from the last or second last layer of Task 1,
    \item \( \mathbf{W}_{task_{2}} \): Weight matrix from the last or second last layer of Task 2.
\end{itemize}

The term \( \|\mathbf{W}_{task_{1}} - \mathbf{W}_{task_{2}}\|_2 \) represents the $L_{2}$-norm (Euclidean distance) between the weight matrices of the two tasks' specific layers, enforcing soft parameter sharing between tasks \cite{rotman-reichart-2019-deep}.

\subsection{Model configurations:}
We conducted all our experiments on a single NVIDIA A100 GPU card.
We presented our considered set of VMLMs and their respective versions in Table \ref{tab:model_version_vmlm}. This set includes both open-source and closed-source source VMLMs. For closed source models like GPT-4, it’s important to note that their weights might change in the future as they are updated and improved. We conducted all our experiments with these models during the period from September, 2024 to November, 2024. We used a default set of hyperparameters for VMLMs using LLaMA-Factory \cite{zheng2024llamafactory} across runs to maintain consistency in the results (see Table \ref{tab:gen_para_vmlm}). 

\begingroup
\renewcommand{\arraystretch}{1} 
\begin{table}[H]
    \centering
    \setkeys{Gin}{keepaspectratio}
    \resizebox*{\columnwidth}{0.99\textheight} {
    \begin{tabular}{l|c}
    \hline
        Parameters & Values \\ \hline \hline
        Learning rate & \{2e-6, 2e-5, 2e-4, 3e-3, 9e-3, 1e-2\} \\ 
        Optimizer & SGD,  AdamW \\ 
        Gamma value (Scheduler) & 0.9, 0.8 \\
        Loss & Weighted CE \\
        Weights (loss) & $ \left[  \frac{ N } { P+N } , \frac{ P } { P+N } \right]$ \\
        Batch size & 16, 32, 64 \\
        Sequence length & 64, 128, 248 \\ 
        Patience (Early stop) & 4 \\
        Regularization strength & \{0, 5e-1, 5e-2, 5e-3, 5e-4\} \\  
        Number of few shots & \{0, 2, 4, 8, 12\} \\ \hline
    \end{tabular}}

    \caption{Model configurations for experiments. Notation: `P' for number of positive sample and `N' for number of negative sample.}
    \label{tab:model_config}
\end{table}
\endgroup

\begingroup
\renewcommand{\arraystretch}{1} 
\begin{table}[H]
    \centering
    \setkeys{Gin}{keepaspectratio}
    \resizebox*{\columnwidth}{0.99\textheight} {
    \begin{tabular}{l|>{\raggedleft\arraybackslash}p{7cm}}
    \hline
        VMLM & Version \\ \hline \hline
        Gemma & \texttt{Telugu-LLM-Labs/Indic-gemma-7b-} \\
        ~ & \texttt{finetuned-sft-Navarasa-2.0} \\
        Aya Expanse & \texttt{CohereForAI/aya-expanse-8b} \\

        Llama-3.1 & \texttt{meta-llama/Llama-3.1-8B-Instruct} \\ 
        GPT-4 & \texttt{ChatGPT} \\ \hline
    \end{tabular}}

    \caption{Model versions in VMLMs. }
    \label{tab:model_version_vmlm}
\end{table}
\endgroup

\begingroup
\renewcommand{\arraystretch}{1} 
\begin{table}[H]
    \centering
    \setkeys{Gin}{keepaspectratio}
    \resizebox*{0.65\columnwidth}{\textheight} {
    \begin{tabular}{l|>{\raggedleft\arraybackslash}p{2.5cm}}
    \hline
        Parameter & Value \\ \hline \hline
        do\_sample & \texttt{true} \\
        temperature & \texttt{0.95} \\
        top\_p & \texttt{0.7} \\ 
        top\_k & \texttt{50} \\ 
        num\_beams & \texttt{1} \\
        max\_length & \texttt{1024} \\
        max\_new\_tokens & \texttt{1024} \\ 
        repetition\_penalty & \texttt{1.0} \\
        length\_penalty & \texttt{1.0} \\
        skip\_special\_tokens & \texttt{true} \\ \hline
    \end{tabular}}

    \caption{Generation parameters in VMLMs. }
    \label{tab:gen_para_vmlm}
\end{table}
\endgroup

\section{Ablation study:}
\label{section:Ablation Study}
\begin{figure*}
    \centering
        \subfigure[Code-mixed humor detection]{\includegraphics[width=0.49\textwidth]{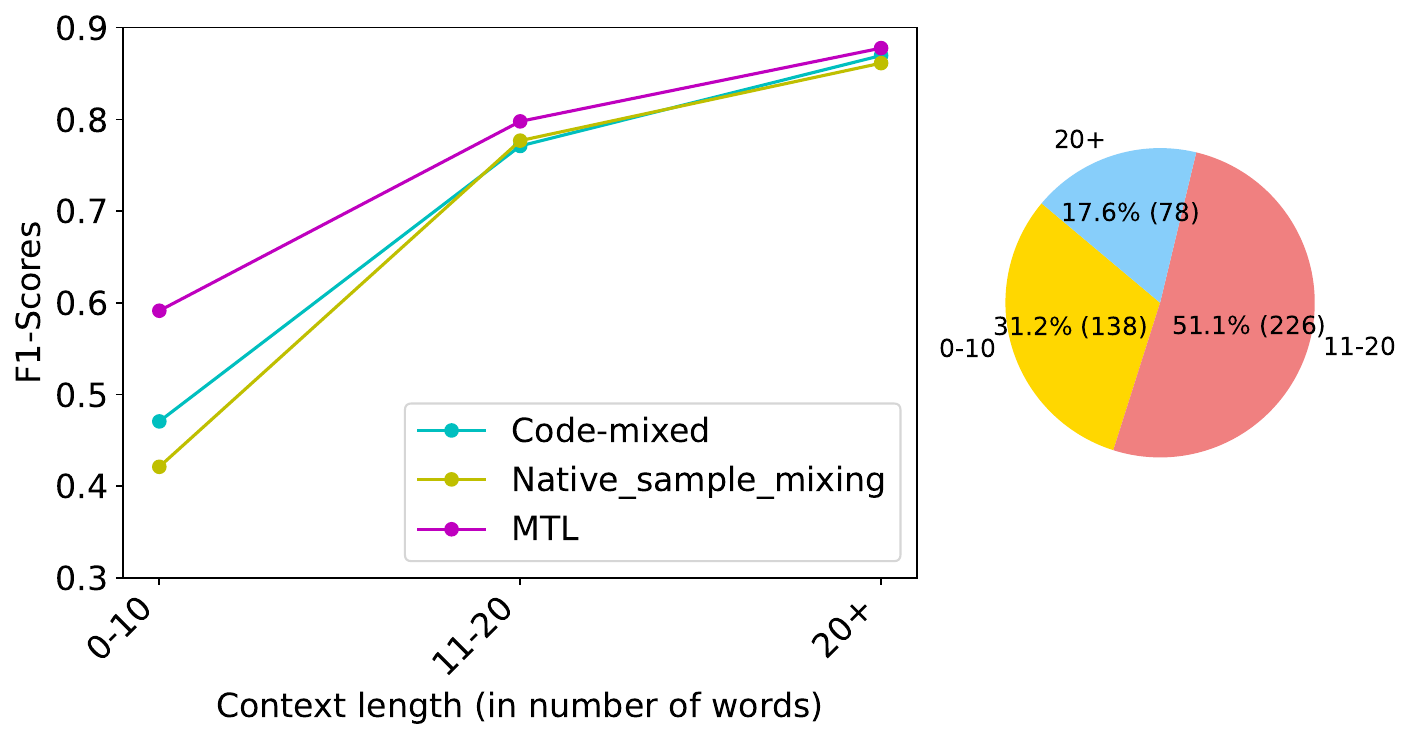}}
        \subfigure[Code-mixed sarcasm detection]{\includegraphics[width=0.49\textwidth]{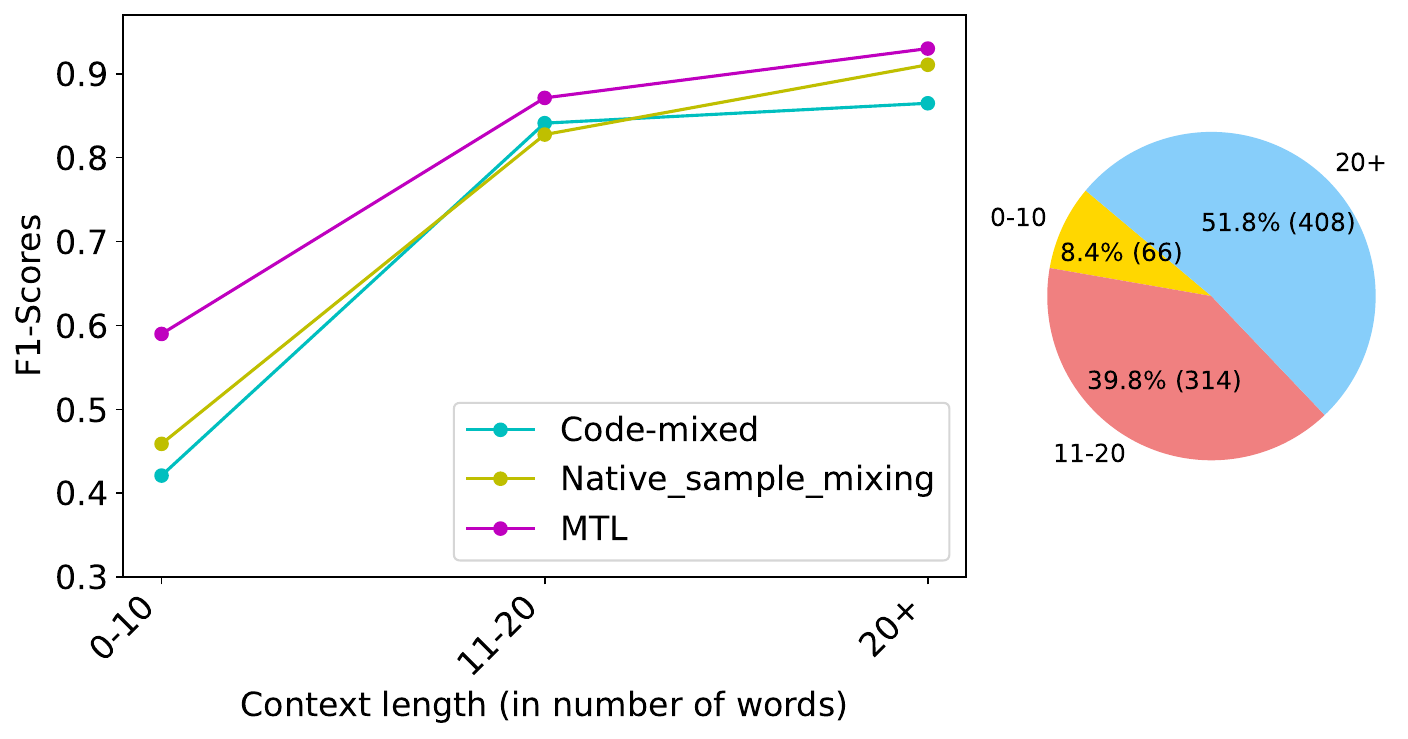}}
    \caption{Performance analysis with increasing context length. Here, the corresponding pie-chart represents the distribution of context length.}
    \label{fig:context_len}
\end{figure*}

In this section, to analyze the role of gating component within the multi-task learning model, we removed it to compare performance and the outcomes are presented in Table \ref{tab:MTL_humor_results}.
Key observations include:

\begin{enumerate}
    \item \textbf{Spelling errors:} For instance, consider the sarcastic statement:  \textit{``@flypigmk uski g**d mein dum hai.. agar kisi aur ke g**d mein nahi hai to uske baap ka kya jaat hai... \#sarcasm with \#g**d''} (\textbf{Gloss:} @flypigmk, he has strong a**... if someone else doesn't have the a**, what does that say about his father's caste... \#sarcasm with \#a**). Here, the MTL model with gating is able to detect the typo error, \textit{`jata'}(\textbf{Gloss:} goes) is misspelled as \textit{`jaat'}(\textbf{Gloss:} caste), however the MTL without gating got confused. 
    
    \item \textbf{Shorter context:} For example, in the humorous sample:  \textit{``Sir @arvindkejriwal AAP karen to chamatkaar, BJP kare to balatkaar.''}(\textbf{Gloss:} Sir @arvindkejriwal, If AAP does it, it's a miracle, if BJP does it, it's a rape.), the gating mechanism overtook the model without gating by detecting humorous contrast using the rhyming words \textit{``chamatkaar''} (\textbf{Gloss:} miracle) and \textit{``balatkaar''} (\textbf{Gloss:} rape) within shorter context. In a similar way, in the sarcastic statement: \textit{``@iamyasaar Flop graphy Ki baat Goti fan ke muh se?? \#Irony''}(\textbf{Gloss:}  @iamyasaar Talking about flop graphy from the mouth of a Goti fan?? \#Irony), gated model was able to detect the ironic situation where a person who is perceived to be a fan of something unsuccessful is commenting on another failure, within such shorter context.
\end{enumerate}

\section{Examination of translated Hindi data:}
\label{appendix:translation_hin}

\begin{figure}[H]
    \centering
        \includegraphics[trim=2.5cm 11.2cm 10.7cm 7.5cm, clip, width=\columnwidth]{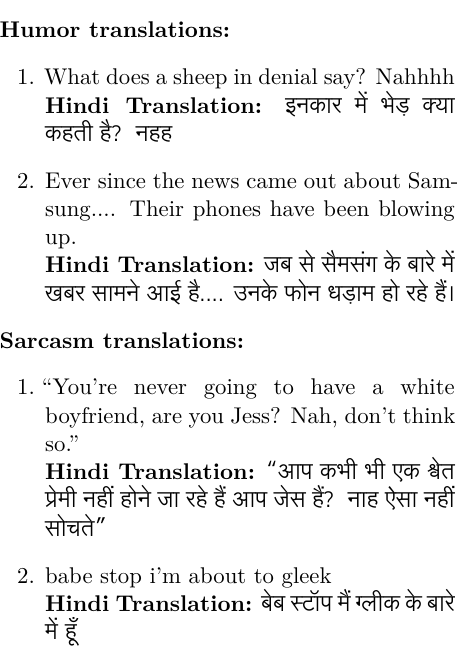}
    \caption{Translated samples of Hindi humor and sarcasm directly from the native English dataset. }
    \label{fig:humor_sarcasm_hindi_trans}
\end{figure}

In this section, we reported our qualitative investigation of code-mixed and translated Hindi samples. This investigation led to two crucial observations. We first observed that most of the code-mixed humour and sarcasm samples are Hindi dominated. Secondly, for many samples humour and sarcasm got lost when they were translated from English samples. To showcase it, we reported some examples of Hindi translation obtained using Google Translate API in Figure \ref{fig:humor_sarcasm_hindi_trans}.
The humor and sarcasm in the English samples often rely on wordplay, puns, ironic and idiomatic expressions that may not have direct equivalents in Hindi. The translated versions attempt a literal translation, losing the subtleties, play on words and cultural context present in the original English samples. In the first humor example, the Hindi translation fails to capture the wordplay of \textit{``denial''} and \textit{``Nahhhh''}(onomatopoeic word used for sheep), resulting in a literal and less humorous translation. In the second humor example, the Hindi translation fails to capture the play on words related to the news about Samsung phones ``blowing up'', as the literal translation does not convey the intended humor. In a similar way for sarcasm examples, the Hindi translation lacks the subtlety and incongruity necessary for sarcasm, as it straightforwardly conveys the situation without emphasizing the ironic tone. This leads to a drop in degree of sarcasm of the translated Hindi version. Thus, the Hindi translations of English (especially more for humor) data samples did not preserve the native cultural context. This analysis emphasizes the need for a more precise context aware translation method. Since, translated Hindi samples didn't preserve the humorous and sarcastic context, we decided to use only English samples for further experiments.

\begingroup
\renewcommand{\arraystretch}{1.15} 
\begin{table*}
    \centering
    \setkeys{Gin}{keepaspectratio}
    \resizebox*{\textwidth}{0.99\textheight} {
    \begin{tabular}{>{\cellcolor{white}}l|>{\cellcolor{white}}cccc>{\cellcolor{white}}c>{\cellcolor{white}}c>{\cellcolor{white}}cccc>{\cellcolor{white}}c>{\cellcolor{white}}c>{\cellcolor{white}}c}
    \Xhline{3\arrayrulewidth}
        \multicolumn{14}{c}{\textbf{\Large{Humor}}}  \\
        \multicolumn{14}{c}{\textbf{\Large{Precision}}}  \\
        \Xhline{3\arrayrulewidth}
        NLD $\rightarrow$ & & \multicolumn{3}{c}{Col} & \multicolumn{3}{c}{POTD} & \multicolumn{3}{c}{HaHa} & \multicolumn{3}{c}{16000} \\  \cmidrule(lr){3-5} \cmidrule(lr){6-8} \cmidrule(lr){9-11} \cmidrule(lr){12-14}
        Model $\downarrow$ & CM & CM+Hi+En & CM+En & CM+Hi & CM+Hi+En & CM+En & CM+Hi & CM+Hi+En & CM+En & CM+Hi & CM+Hi+En & CM+En & CM+Hi \\ \hline \hline
        \rowcolor{gray!10}
        NB & 0.72 & \textbf{0.78} & 0.74 & \textbf{0.77} & \textbf{0.79} & 0.74 & 0.79 & \textbf{0.78} & 0.75 & \textbf{0.79} & 0.76 & 0.78 & 0.73 \\ 
        \rowcolor{gray!10}
        RF & 0.72 & 0.72 & 0.73 & 0.74 & 0.74 & 0.74 & 0.75 & 0.73 & 0.73 & 0.75 & 0.74 & 0.73 & 0.72 \\ 
        \rowcolor{gray!10}
        SVM & 0.73 & 0.73 & 0.70 & 0.71 & 0.73 & 0.70 & 0.74 & 0.72 & 0.74 & 0.73 & 0.72 & 0.72 & 0.74 \\ 
        \rowcolor{gray!10}
        mBERT & \textbf{0.78} & 0.77 & 0.75 & 0.71 & 0.74 & \textbf{0.80} & 0.77 & 0.73 & 0.76 & 0.78 & \textbf{0.78} & \textbf{0.79} & 0.73 \\ 
        \rowcolor{gray!10}
        XLM-R & 0.73 & 0.75 & 0.75 & 0.74 & 0.77 & 0.77 & 0.75 & 0.76 & 0.78 & 0.77 & 0.76 & 0.78 & 0.76 \\ 
        \rowcolor{gray!10}
        MuRIL & 0.77 & 0.76 & 0.76 & \textbf{0.77} & 0.75 & 0.75 & \textbf{0.80} & \textbf{0.78} & \textcolor{blue}{\textbf{0.81}} & 0.78 & 0.77 & 0.78 & \textbf{0.80} \\
        \rowcolor{gray!10}
        IndicBERT & 0.70 & \textbf{0.78} & \textbf{0.77} & 0.74 & \textbf{0.79} & 0.76 & 0.79 & 0.74 & 0.80 & 0.77 & 0.74 & 0.77 & 0.76 \\ \hline

    \end{tabular}}

    \vspace{1mm}

    \setkeys{Gin}{keepaspectratio}
    \resizebox*{\textwidth}{0.99\textheight} {
    \begin{tabular}{>{\cellcolor{white}}l|>{\cellcolor{white}}cccc>{\cellcolor{white}}c>{\cellcolor{white}}c>{\cellcolor{white}}cccc>{\cellcolor{white}}c>{\cellcolor{white}}c>{\cellcolor{white}}c}
    \Xhline{3\arrayrulewidth}
        \multicolumn{14}{c}{\textbf{\Large{Recall}}}  \\
        \Xhline{3\arrayrulewidth}
        NLD $\rightarrow$ & & \multicolumn{3}{c}{Col} & \multicolumn{3}{c}{POTD} & \multicolumn{3}{c}{HaHa} & \multicolumn{3}{c}{16000} \\  \cmidrule(lr){3-5} \cmidrule(lr){6-8} \cmidrule(lr){9-11} \cmidrule(lr){12-14}
        Model $\downarrow$ & CM & CM+Hi+En & CM+En & CM+Hi & CM+Hi+En & CM+En & CM+Hi & CM+Hi+En & CM+En & CM+Hi & CM+Hi+En & CM+En & CM+Hi \\ \hline \hline
        \rowcolor{gray!10}
        NB & 0.78 & 0.72 & \textbf{0.76} & 0.73 & 0.71 & 0.76 & 0.71 & 0.70 & 0.74 & 0.71 & 0.74 & 0.72 & \textbf{0.79} \\ 
        \rowcolor{gray!10}
        RF & 0.72 & 0.67 & 0.68 & 0.65 & 0.66 & 0.70 & 0.67 & 0.70 & 0.68 & 0.69 & 0.65 & 0.65 & 0.64 \\ 
        \rowcolor{gray!10}
        SVM & 0.69 & 0.64 & 0.60 & 0.63 & 0.65 & 0.70 & 0.65 & 0.67 & 0.65 & 0.65 & 0.64 & 0.67 & 0.65 \\ 
        \rowcolor{gray!10}
        mBERT & 0.78 & 0.70 & 0.70 & 0.61 & 0.69 & 0.72 & \textbf{0.73} & 0.64 & \textbf{0.76} & 0.74 & 0.70 & 0.71 & 0.68 \\ 
        \rowcolor{gray!10}
        XLM-R & 0.77 & 0.70 & 0.71 & 0.72 & 0.69 & 0.73 & 0.71 & 0.72 & 0.73 & 0.70 & \textbf{0.76} & 0.74 & 0.68 \\ 
        \rowcolor{gray!10}
        MuRIL & 0.73 & \textbf{0.83} & 0.68 & \textbf{0.75} & 0.71 & \textbf{0.83} & 0.76 & 0.78 & 0.73 & \textbf{0.78} & \textbf{0.76} & \textbf{0.77} & 0.76 \\
        \rowcolor{gray!10}
        IndicBERT & \textbf{0.79} & 0.75 & 0.68 & 0.69 & \textbf{0.75} & 0.77 & 0.71 & \textcolor{blue}{\textbf{0.84}} & 0.72 & 0.73 & 0.75 & 0.69 & 0.76 \\ \hline

    \end{tabular}}

    \vspace{3mm}

    \setkeys{Gin}{keepaspectratio}
    \resizebox*{0.79\textwidth}{0.99\textheight} {
    \begin{tabular}{>{\cellcolor{white}}l|>{\cellcolor{white}}cccc>{\cellcolor{white}}c>{\cellcolor{white}}c>{\cellcolor{white}}cccc}
    \Xhline{3\arrayrulewidth}
        \multicolumn{11}{c}{\textbf{\Large{Sarcasm}}}  \\
        \multicolumn{11}{c}{\textbf{\Large{Precision}}}  \\
        \Xhline{3\arrayrulewidth}
        NLD $\rightarrow$ & & \multicolumn{3}{c}{NHD} & \multicolumn{3}{c}{iSarc} & \multicolumn{3}{c}{SC-V2} \\  \cmidrule(lr){3-5} \cmidrule(lr){6-8} \cmidrule(lr){9-11} 
        Model $\downarrow$ & CM & CM+Hi+En & CM+En & CM+Hi & CM+Hi+En & CM+En & CM+Hi & CM+Hi+En & CM+En & CM+Hi \\ \hline \hline
        \rowcolor{gray!10}
        NB & 0.71 & 0.49 & 0.38 & 0.42 & 0.56 & 0.47 & 0.50 & 0.34 & 0.35 & 0.46 \\ 
        \rowcolor{gray!10}
        RF & 0.84 & 0.50 & 0.60 & 0.57 & 0.52 & 0.56 & 0.58 & 0.53 & 0.61 & 0.79 \\ 
        \rowcolor{gray!10}
        SVM & 0.73 & 0.57 & 0.60 & 0.76 & 0.62 & 0.63 & 0.71 & 0.82 & 0.76 & 0.72 \\ 
        \rowcolor{gray!10}
        mBERT & \textcolor{blue}{\textbf{0.93}} & 0.83 & 0.83 & 0.76 & 0.82 & 0.84 & 0.80 & 0.82 & \textbf{0.85} & 0.82 \\ 
        \rowcolor{gray!10}
        XLM-R & 0.83 & 0.78 & 0.83 & 0.82 & 0.85 & 0.82 & 0.83 & 0.79 & 0.82 & \textbf{0.85} \\ 
        \rowcolor{gray!10}
        MuRIL & 0.84 & \textbf{0.87} & \textbf{0.88} & 0.83 & \textbf{0.87} & \textbf{0.89} & 0.86 & \textbf{0.83} & \textbf{0.85} & 0.84 \\ 
        \rowcolor{gray!10}
        IndicBERT & 0.82 & 0.86 & 0.85 & \textbf{0.86} & 0.81 & 0.88 & \textbf{0.89} & 0.82 & 0.84 & 0.83 \\ \hline

    \end{tabular}}

    \vspace{1mm}

    \setkeys{Gin}{keepaspectratio}
    \resizebox*{0.79\textwidth}{0.99\textheight} {
    \begin{tabular}{>{\cellcolor{white}}l|>{\cellcolor{white}}cccc>{\cellcolor{white}}c>{\cellcolor{white}}c>{\cellcolor{white}}cccc}
    \Xhline{3\arrayrulewidth}
        \multicolumn{11}{c}{\textbf{\Large{Recall}}}  \\
        \Xhline{3\arrayrulewidth}
        NLD $\rightarrow$ & & \multicolumn{3}{c}{NHD} & \multicolumn{3}{c}{iSarc} & \multicolumn{3}{c}{SC-V2} \\  \cmidrule(lr){3-5} \cmidrule(lr){6-8} \cmidrule(lr){9-11} 
        Model $\downarrow$ & CM & CM+Hi+En & CM+En & CM+Hi & CM+Hi+En & CM+En & CM+Hi & CM+Hi+En & CM+En & CM+Hi \\ \hline \hline
        \rowcolor{gray!10}
        NB & 0.80 & 0.29 & 0.36 & 0.40 & 0.31 & 0.39 & 0.37 & 0.33 & 0.34 & 0.33 \\ 
        \rowcolor{gray!10}
        RF & 0.59 & 0.38 & 0.47 & 0.63 & 0.51 & 0.59 & 0.55 & 0.44 & 0.60 & 0.50 \\ 
        \rowcolor{gray!10}
        SVM & 0.79 & 0.61 & 0.59 & 0.72 & 0.67 & 0.66 & 0.71 & 0.64 & 0.63 & 0.73 \\ 
        \rowcolor{gray!10}
        mBERT & 0.70 & 0.83 & 0.72 & \textbf{0.85} & 0.77 & 0.78 & 0.75 & 0.75 & 0.79 & 0.88 \\ 
        \rowcolor{gray!10}
        XLM-R & 0.80 & 0.85 & 0.76 & 0.83 & 0.78 & 0.80 & 0.80 & 0.84 & 0.83 & 0.79 \\ 
        \rowcolor{gray!10}
        MuRIL & \textbf{0.82} & 0.85 & \textbf{0.89} & 0.81 & \textbf{0.83} & 0.82 & \textbf{0.84} & \textcolor{blue}{\textbf{0.91}} & \textbf{0.85} & \textbf{0.90} \\ 
        \rowcolor{gray!10}
        IndicBERT & 0.80 & \textbf{0.86} & 0.87 & 0.84 & 0.81 & \textbf{0.88} & 0.78 & 0.84 & 0.82 & 0.85 \\ \hline

    \end{tabular}}

    \caption{Precision and recall scores for humor (first and second table) and sarcasm (third and fourth table) detection using native sample mixing. Notation: NLD for native language dataset, CM for code-mixed. The scores of best-performing models for the individual training scenarios are marked in bold and highest in the respective task is marked in \textcolor{blue}{blue}.}
    \label{tab:results_native_humor_precall}
\end{table*}
\endgroup
\begingroup
\renewcommand{\arraystretch}{1.15} 
\begin{table*}
    \centering
    \setkeys{Gin}{keepaspectratio}
    \resizebox*{\textwidth}{\textheight} {
    \begin{tabular}{ll|cccccc|cccccc}
    \Xhline{3\arrayrulewidth}
        \multicolumn{14}{c}{\textbf{\Large{Humor}}}  \\
        \Xhline{3\arrayrulewidth}
        ~ & ~ & \multicolumn{6}{c|}{\textbf{Precision}} & \multicolumn{6}{c}{\textbf{Recall}}  \\  \hline
        \multicolumn{2}{c|}{NLD : Col} &  \multicolumn{2}{c}{$\text{mBERT}_{MTL}$}  & \multicolumn{2}{c}{$\text{XLM-R}_{MTL}$} & \multicolumn{2}{c}{$\text{MuRIL}_{MTL}$} &  \multicolumn{2}{c}{$\text{mBERT}_{MTL}$}  & \multicolumn{2}{c}{$\text{XLM-R}_{MTL}$} & \multicolumn{2}{c}{$\text{MuRIL}_{MTL}$}  \\  \cmidrule(lr){1-2} \cmidrule(lr){3-4} \cmidrule(lr){5-6} \cmidrule(lr){7-8} \cmidrule(lr){9-10} \cmidrule(lr){11-12} \cmidrule(lr){13-14}
        Hate & Sarcasm  & Gate & w/o Gate & Gate & w/o Gate & Gate & w/o Gate & Gate & w/o Gate & Gate & w/o Gate & Gate & w/o Gate \\ \hline \hline
        \checkboxempty & $\checkboxcmark_{NHD}$ & 0.72 & 0.75 & \textbf{0.82} & 0.78 & 0.77 & 0.72 & 0.63 & 0.77 & 0.67 & \textbf{0.78} & 0.77 & 0.68\\
        \checkboxempty & $\checkboxcmark_{iSarc}$ & 0.74 & \textbf{0.79} & \textbf{0.79} & 0.77 & 0.76 & 0.73 & 0.65 & 0.77 & \textbf{0.81} & 0.77 & 0.75 & 0.72 \\
        \checkboxempty & $\checkboxcmark_{SC-V2}$ & 0.70 & 0.76 & 0.76 & 0.75 & \textbf{0.78} & 0.77 & 0.71 & 0.74 & \textbf{0.76} & \textbf{0.76} & 0.74 & \textbf{0.76} \\
        $\checkboxcmark$ & \checkboxempty & 0.74 & 0.76 & 0.76 & \textbf{0.77} & \textbf{0.77} & \textbf{0.77} & \textbf{0.79} & 0.72 & 0.71 & 0.75 & 0.77 & 0.75 \\
        $\checkboxcmark$ & $\checkboxcmark_{NHD}$ & \textbf{0.79} & 0.77 & 0.76 & 0.77 & 0.78 & 0.76 & 0.75 & \textbf{0.81} & \textbf{0.81} & 0.77 & 0.79 & 0.78 \\
        $\checkboxcmark$ & $\checkboxcmark_{iSarc}$ & \textbf{0.79} & 0.78 & 0.77 & \textbf{0.79} & 0.74 & 0.77 & 0.78 & 0.77 & \textbf{0.81} & 0.79 & 0.79 & 0.76 \\
        $\checkboxcmark$ & $\checkboxcmark_{SC-V2}$ & \textbf{0.78} & 0.77 & 0.74 & 0.76 & \textbf{0.78} & 0.77 & 0.75 & \textbf{0.79} & 0.78 & 0.78 & 0.77 & 0.75 \\ \hline
        
        \multicolumn{2}{c|}{NLD : POTD}  & & & & & \\ \cmidrule(lr){1-2}
        \checkboxempty & $\checkboxcmark_{NHD}$ & 0.73 & 0.77 & \textbf{0.79} & \textbf{0.79} & 0.75 & 0.74 & 0.69 & 0.80 & 0.79 & \textbf{0.82} & 0.72 & 0.71 \\
        \checkboxempty & $\checkboxcmark_{iSarc}$ & 0.74 & \textbf{0.78} & \textbf{0.78} & 0.77 & 0.77 & 0.77 & \textbf{0.81} & \textbf{0.81} & 0.78 & 0.80 & 0.76 & 0.80 \\
        \checkboxempty & $\checkboxcmark_{SC-V2}$ & 0.74 & \textbf{0.78} & 0.77 & 0.77 & \textbf{0.78} & 0.75 & 0.71 & 0.79 & 0.73 & \textbf{0.81} & 0.76 & 0.72 \\
        $\checkboxcmark$ & \checkboxempty & 0.77 & 0.76 & \textbf{0.79} & 0.77 & 0.75 & 0.73 & 0.68 & \textbf{0.80} & 0.74 & \textbf{0.80} & 0.67 & 0.70 \\
        $\checkboxcmark$ & $\checkboxcmark_{NHD}$ & \textbf{0.78} & 0.77 & \textbf{0.78} & 0.77 & 0.77 & 0.77 & 0.80 & 0.80 & \textbf{0.82} & 0.80 & 0.75 & 0.73 \\
        $\checkboxcmark$ & $\checkboxcmark_{iSarc}$ & 0.75 & \textbf{0.79} & 0.77 & 0.77 & 0.74 & 0.77 & \textcolor{blue}{\textbf{0.85}} & 0.77 & 0.80 & 0.77 & 0.81 & 0.79 \\
        $\checkboxcmark$ & $\checkboxcmark_{SC-V2}$ & 0.75 & 0.78 & \textbf{0.79} & 0.72 & 0.75 & 0.76 & \textcolor{blue}{\textbf{0.85}} & 0.81 & 0.81 & 0.74 & 0.70 & 0.74 \\ \hline
        
        \multicolumn{2}{c|}{NLD : HaHa}  & & & & & \\ \cmidrule(lr){1-2}
        \checkboxempty & $\checkboxcmark_{NHD}$ & \textbf{0.80} & \textbf{0.80} & 0.78 & 0.77 & 0.74 & 0.78 & 0.74 & 0.77 & \textbf{0.82} & 0.79 & 0.71 & 0.75 \\
        \checkboxempty & $\checkboxcmark_{iSarc}$ & 0.76 & \textbf{0.80} & 0.78 & 0.78 & 0.76 & 0.77 & 0.77 & 0.79 & \textbf{0.81} & 0.76 & 0.68 & 0.74 \\
        \checkboxempty & $\checkboxcmark_{SC-V2}$ & 0.78 & 0.79 & \textbf{0.80} & 0.77 & 0.78 & 0.78 & \textbf{0.80} & 0.76 & 0.76 & 0.79 & 0.70 & 0.74 \\
        $\checkboxcmark$ & \checkboxempty & 0.77 & 0.74 & \textbf{0.80} & 0.79 & 0.78 & 0.77 & 0.70 & 0.70 & 0.77 & 0.75 & \textbf{0.82} & 0.75 \\
        $\checkboxcmark$ & $\checkboxcmark_{NHD}$ & \textbf{0.79} & 0.77 & 0.77 & 0.78 & 0.77 & 0.77 & 0.75 & 0.80 & \textbf{0.81} & 0.80 & 0.80 & 0.75 \\
        $\checkboxcmark$ & $\checkboxcmark_{iSarc}$ & \textbf{0.80} & 0.78 & 0.79 & 0.78 & 0.79 & 0.78 & 0.77 & \textbf{0.82} & 0.77 & 0.76 & 0.78 & 0.79 \\
        $\checkboxcmark$ & $\checkboxcmark_{SC-V2}$ & 0.77 & 0.78 & 0.77 & 0.79 & \textcolor{blue}{\textbf{0.81}} & 0.79 & 0.80 & 0.81 & 0.79 & 0.76 & \textbf{0.83} & 0.82 \\ \hline
        
        \multicolumn{2}{c|}{NLD : 16000}  & & & & & \\ \cmidrule(lr){1-2}
        \checkboxempty & $\checkboxcmark_{NHD}$ & 0.77 & 0.71 & \textbf{0.78} & \textbf{0.78} & 0.74 & \textbf{0.78} & 0.75 & 0.67 & 0.79 & \textbf{0.82} & 0.68 & 0.75 \\
        \checkboxempty & $\checkboxcmark_{iSarc}$ & 0.76 & 0.78 & 0.78 & \textbf{0.79} & 0.77 & 0.75 & 0.77 & 0.77 & \textbf{0.81} & 0.77 & 0.79 & 0.70 \\
        \checkboxempty & $\checkboxcmark_{SC-V2}$ & 0.78 & 0.76 & \textbf{0.79} & 0.78 & 0.78 & 0.73 & \textbf{0.80} & 0.70 & 0.78 & \textbf{0.80} & 0.72 & 0.68 \\
        $\checkboxcmark$ & \checkboxempty & \textbf{0.79} & 0.77 & \textbf{0.79} & 0.76 & 0.78 & 0.75 & 0.78 & \textbf{0.79} & 0.77 & 0.78 & 0.77 & 0.71 \\
        $\checkboxcmark$ & $\checkboxcmark_{NHD}$ & 0.78 & 0.78 & \textbf{0.79} & 0.78 & \textbf{0.79} & 0.78 & 0.75 & 0.80 & 0.77 & \textbf{0.81} & 0.78 & 0.75 \\
        $\checkboxcmark$ & $\checkboxcmark_{iSarc}$ & 0.78 & 0.78 & \textbf{0.80} & 0.79 & \textbf{0.80} & 0.79 & \textbf{0.79} & 0.78 & \textbf{0.79} & 0.76 & 0.78 & 0.77 \\
        $\checkboxcmark$ & $\checkboxcmark_{SC-V2}$ & \textbf{0.79} & 0.78 & 0.78 & \textbf{0.79} & \textbf{0.79} & \textbf{0.79} & 0.77 & \textbf{0.81} & \textbf{0.81} & 0.77 & \textbf{0.81} & 0.78 \\ \hline
    \end{tabular}}

    \caption{Precision and recall scores of MTL framework and ablation study for humor detection. Notation: `NLD' for Native Language Dataset. The scores of best-performing models for the individual training scenarios are marked in bold and highest in the respective task is marked in \textcolor{blue}{blue}.}
    \label{tab:MTL_humor_results_precall}
\end{table*}
\endgroup

\begingroup
\renewcommand{\arraystretch}{1.15} 
\begin{table*}
    \centering
    \setkeys{Gin}{keepaspectratio}
    \resizebox*{\textwidth}{\textheight} {
    \begin{tabular}{ll|cccccc|cccccc}
    \Xhline{3\arrayrulewidth}
        \multicolumn{14}{c}{\textbf{\Large{Sarcasm}}}  \\
        \Xhline{3\arrayrulewidth}
        ~ & ~ & \multicolumn{6}{c|}{\textbf{Precision}} & \multicolumn{6}{c}{\textbf{Recall}}  \\  \hline
        \multicolumn{2}{c|}{NLD : NHD} &  \multicolumn{2}{c}{$\text{mBERT}_{MTL}$}  & \multicolumn{2}{c}{$\text{XLM-R}_{MTL}$} & \multicolumn{2}{c}{$\text{MuRIL}_{MTL}$} &  \multicolumn{2}{c}{$\text{mBERT}_{MTL}$}  & \multicolumn{2}{c}{$\text{XLM-R}_{MTL}$} & \multicolumn{2}{c}{$\text{MuRIL}_{MTL}$}  \\  \cmidrule(lr){1-2} \cmidrule(lr){3-4} \cmidrule(lr){5-6} \cmidrule(lr){7-8} \cmidrule(lr){9-10} \cmidrule(lr){11-12} \cmidrule(lr){13-14}
        Hate & Humor  & Gate & w/o Gate & Gate & w/o Gate & Gate & w/o Gate & Gate & w/o Gate & Gate & w/o Gate & Gate & w/o Gate \\ \hline \hline
        \checkboxempty & $\checkboxcmark_{Col}$ & 0.82 & \textbf{0.83} & \textbf{0.83} & 0.81 & 0.82 & 0.78 & 0.82 & 0.83 & \textbf{0.87} & 0.83 & 0.83 & 0.78  \\
        \checkboxempty & $\checkboxcmark_{POTD}$ & 0.82 & \textbf{0.85} & 0.82 & 0.83 & 0.83 & 0.79 & 0.85 & 0.81 & \textbf{0.88} & 0.82 & 0.81 & 0.74 \\
        \checkboxempty & $\checkboxcmark_{HaHa}$ & 0.83 & \textbf{0.85} & 0.83 & 0.83 & 0.83 & 0.80 & 0.85 & 0.81 & \textbf{0.86} & 0.81 & 0.84 & 0.77 \\
        \checkboxempty & $\checkboxcmark_{16000}$ & 0.84 & 0.84 & \textcolor{blue}{\textbf{0.88}} & 0.83 & 0.84 & 0.80 & 0.79 & 0.82 & \textbf{0.90} & 0.83 & 0.88 & 0.78 \\
        \checkboxcmark & $\checkboxempty$ & 0.81 & \textbf{0.85} & 0.83 & 0.83 & 0.84 & 0.79 & 0.81 & 0.85 & \textbf{0.88} & 0.81 & 0.87 & 0.78 \\
        \checkboxcmark & $\checkboxcmark_{Col}$ & 0.83 & 0.82 & 0.83 & \textbf{0.87} & \textbf{0.87} & 0.84 & 0.83 & 0.88 & 0.84 & \textbf{0.90} & 0.84 & 0.83 \\
        \checkboxcmark & $\checkboxcmark_{POTD}$ & 0.83 & 0.86 & 0.85 & \textbf{0.87} & 0.83 & 0.82 & 0.86 & 0.79 & \textcolor{blue}{\textbf{0.92}} & 0.87 & 0.84 & 0.82 \\
        \checkboxcmark & $\checkboxcmark_{HaHa}$ & 0.80 & 0.84 & \textcolor{blue}{\textbf{0.88}} & 0.86 & 0.81 & 0.84 & 0.87 & 0.83 & \textbf{0.88} & \textbf{0.88} & 0.81 & 0.78 \\
        \checkboxcmark & $\checkboxcmark_{16000}$ & 0.84 & 0.84 & 0.83 & 0.82 & \textbf{0.87} & 0.80 & 0.86 & 0.86 & \textbf{0.90} & 0.82 & 0.87 & 0.82 \\ \hline
        
        \multicolumn{2}{c|}{NLD : iSarc}  & & & & & \\ \cmidrule(lr){1-2}
        \checkboxempty & $\checkboxcmark_{Col}$ & 0.80 & 0.83 & 0.80 & 0.82 & \textcolor{blue}{\textbf{0.88}} & 0.77 & 0.83 & 0.83 & \textbf{0.86} & 0.80 & 0.85 & 0.79 \\
        \checkboxempty & $\checkboxcmark_{POTD}$ & 0.80 & \textbf{0.84} & 0.83 & 0.82 & 0.83 & 0.79 & \textbf{0.85} & 0.82 & 0.83 & 0.80 & 0.80 & 0.80 \\
        \checkboxempty & $\checkboxcmark_{HaHa}$ & \textbf{0.86} & 0.85 & 0.83 & 0.84 & 0.83 & 0.80 & 0.81 & 0.80 & \textbf{0.85} & \textbf{0.85} & \textbf{0.85} & 0.77 \\
        \checkboxempty & $\checkboxcmark_{16000}$ & 0.82 & 0.83 & 0.80 & 0.81 & \textbf{0.85} & 0.80 & 0.79 & \textbf{0.87} & 0.84 & 0.80 & 0.82 & 0.78 \\
        \checkboxcmark & $\checkboxempty$ & 0.82 & 0.81 & 0.82 & 0.83 & \textbf{0.85} & 0.76 & 0.81 & \textbf{0.85} & 0.83 & 0.79 & \textbf{0.85} & 0.73 \\
        \checkboxcmark & $\checkboxcmark_{Col}$ & 0.82 & 0.81 & 0.82 & 0.82 & \textbf{0.86} & 0.80 & 0.80 & 0.81 & \textbf{0.88} & 0.81 & 0.83 & 0.78 \\
        \checkboxcmark & $\checkboxcmark_{POTD}$ & 0.83 & 0.80 & \textbf{0.87} & 0.84 & 0.82 & 0.80 & 0.81 & 0.73 & \textbf{0.85} & 0.81 & 0.81 & 0.78 \\
        \checkboxcmark & $\checkboxcmark_{HaHa}$ & 0.81 & 0.82 & \textcolor{blue}{\textbf{0.88}} & 0.86 & 0.83 & 0.81 & 0.82 & 0.79 & 0.84 & 0.83 & \textbf{0.86} & 0.79 \\
        \checkboxcmark & $\checkboxcmark_{16000}$ & 0.82 & 0.81 & \textbf{0.86} & 0.83 & 0.82 & 0.80 & 0.82 & 0.81 & \textbf{0.87} & 0.84 & 0.82 & 0.78 \\ \hline

        \multicolumn{2}{c|}{NLD : SC-V2}  & & & & & \\ \cmidrule(lr){1-2}
        \checkboxempty & $\checkboxcmark_{Col}$ & 0.84 & \textbf{0.85} & 0.83 & 0.82 & 0.83 & 0.83 & \textbf{0.84} & 0.82 & 0.81 & 0.82 & 0.81 & 0.82 \\
        \checkboxempty & $\checkboxcmark_{POTD}$ & 0.83 & 0.83 & \textbf{0.86} & 0.82 & 0.84 & 0.83 & 0.82 & 0.82 & \textbf{0.83} & 0.81 & 0.82 & 0.82 \\
        \checkboxempty & $\checkboxcmark_{HaHa}$ & 0.84 & 0.85 & 0.82 & 0.84 & \textbf{0.86} & 0.83 & \textbf{0.87} & 0.81 & 0.86 & 0.85 & 0.84 & 0.82 \\
        \checkboxempty & $\checkboxcmark_{16000}$ & 0.82 & 0.83 & \textbf{0.84} & \textbf{0.84} & 0.82 & 0.83 & 0.80 & 0.82 & 0.82 & \textbf{0.84} & \textbf{0.84} & 0.80 \\
        \checkboxcmark & $\checkboxempty$ & 0.83 & 0.83 & 0.84 & 0.83 & \textbf{0.87} & 0.81 & 0.82 & 0.83 & \textbf{0.87} & 0.80 & 0.82 & 0.81 \\
        \checkboxcmark & $\checkboxcmark_{Col}$ & 0.83 & 0.83 & \textbf{0.85} & \textbf{0.85} & \textbf{0.85} & 0.81 & 0.80 & 0.80 & 0.85 & \textbf{0.88} & 0.85 & 0.79\\
        \checkboxcmark & $\checkboxcmark_{POTD}$ & 0.82 & 0.83 & \textbf{0.85} & \textbf{0.85} & 0.84 & 0.84 & \textbf{0.86} & 0.85 & 0.83 & 0.81 & 0.85 & 0.81 \\
        \checkboxcmark & $\checkboxcmark_{HaHa}$ & 0.85 & \textcolor{blue}{\textbf{0.88}} & 0.84 & 0.87 & 0.85 & 0.82 & 0.84 & 0.82 & 0.87 & \textbf{0.88} & \textbf{0.88} & 0.79 \\
        \checkboxcmark & $\checkboxcmark_{16000}$ & 0.83 & 0.83 & \textbf{0.86} & 0.84 & 0.83 & 0.83 & 0.81 & 0.80 & \textbf{0.89} & 0.82 & 0.85 & 0.80 \\ \hline
    \end{tabular}}

    \caption{Precision and recall scores of MTL framework and ablation study for sarcasm detection. Notation: `NLD' for Native Language Dataset. The scores of best-performing models for the individual training scenarios are marked in bold and highest in the respective task is marked in \textcolor{blue}{blue}.}
    \label{tab:MTL_sarcasm_results_precall}
\end{table*}
\endgroup
%
\begingroup
\renewcommand{\arraystretch}{1.08} 
\begin{table*}
    \centering
    \setkeys{Gin}{keepaspectratio}
    \resizebox*{0.9\textwidth}{\textwidth} {
    \begin{tabular}{l|ccccc|ccccc}
    \Xhline{3\arrayrulewidth}
        \multicolumn{11}{c}{\textbf{\Large{Humor}}}  \\
        \Xhline{3\arrayrulewidth}
        ~ & \multicolumn{5}{c|}{\textbf{Precision}} & \multicolumn{5}{c}{\textbf{Recall}}  \\  \hline
        Model & CM & Col & POTD & HaHa & 16000 & CM & Col & POTD & HaHa & 16000 \\ \hline \hline
        Gemma & 0.06 & 0.05 & 0.05 & 0.06 & 0.06 & 0.66 & 0.66 & 0.66 & 0.67 & 0.66  \\
        Aya Expanse & 0.63 & \textbf{0.64} & 0.63 & \textbf{0.60} & \textbf{0.63} & 0.85 & 0.85 & 0.85 & 0.84 & 0.85  \\
        Llama-3.1 & 0.61 & 0.62 & \textcolor{blue}{\textbf{0.65}} & 0.59 & 0.62 & \textcolor{blue}{\textbf{0.94}} & \textbf{0.86} & \textbf{0.86} & \textbf{0.86} & \textbf{0.86}  \\
        GPT-4 & \textcolor{blue}{\textbf{0.65}} & 0.46 & 0.51 & 0.45 & 0.44 & 0.86 & 0.80 & 0.81 & 0.79 & 0.79 \\ \hline

    \end{tabular}
    }

    \vspace{3mm}

    \setkeys{Gin}{keepaspectratio}
    \resizebox*{0.73\textwidth}{\textwidth} {
    \begin{tabular}{l|cccc|cccc}
    \Xhline{3\arrayrulewidth}
        \multicolumn{9}{c}{\textbf{\Large{Sarcasm}}}  \\
        \Xhline{3\arrayrulewidth}
        ~ & \multicolumn{4}{c|}{\textbf{Precision}} & \multicolumn{4}{c}{\textbf{Recall}}  \\  \hline
        Model & CM & NHD & iSarc & SC-V2 & CM & NHD & iSarc & SC-V2 \\ \hline \hline
        Gemma & 0.26 & 0.08 & 0.29 & 0.37 & 0.74 & 0.67 & 0.76 & 0.78 \\
        Aya Expanse &  0.12 & 0.12 & 0.15 & 0.13 & \textcolor{blue}{\textbf{0.94}} & \textcolor{blue}{\textbf{0.94}} & 0.93 & \textbf{0.93}  \\
        Llama-3.1 & 0.12 & \textbf{0.31} & \textbf{0.31} & 0.18 & \textcolor{blue}{\textbf{0.94}} & 0.78 & 0.80 & 0.70  \\ 
        GPT-4 & \textcolor{blue}{\textbf{0.65}} & 0.12 & 0.14 & \textbf{0.39} & \textcolor{blue}{\textbf{0.94}} & 0.67 & \textcolor{blue}{\textbf{0.94}} & 0.76 \\ \hline

    \end{tabular}
    }

    \caption{Precision and recall scores of our experiment evaluating the impact of in-context learning on VMLMs. Notation: CM for code-mixed. The scores of best-performing models for the individual training scenarios are marked in bold and highest in the respective task is marked in \textcolor{blue}{blue}.}
    \label{tab:results_vmlm_precall}
\end{table*}
\endgroup

%
\begingroup
\renewcommand{\arraystretch}{1.08} 
\begin{table*}
    \centering
    \setkeys{Gin}{keepaspectratio}
    \resizebox*{\textwidth}{\textwidth} {
    \begin{tabular}{l|ccccc|ccccc}
    \Xhline{3\arrayrulewidth}
        \multicolumn{11}{c}{\textbf{\Large{Humor}}}  \\
        \Xhline{3\arrayrulewidth}
        ~ & \multicolumn{5}{c|}{\textbf{Precision}} & \multicolumn{5}{c}{\textbf{Recall}}  \\  \hline
        Model & CM & CM+Col & CM+POTD & CM+HaHa & CM+16000 & CM & CM+Col & CM+POTD & CM+HaHa & CM+16000 \\ \hline \hline
        Gemma & \textcolor{blue}{\textbf{0.71}} & 0.52 & 0.64 & 0.35 & 0.62  & 0.87 & 0.78 & \textcolor{blue}{\textbf{0.96}} & 0.66 & \textbf{0.94}  \\
        Aya Expanse & 0.60 & \textbf{0.66} & \textbf{0.67} & \textbf{0.66} & 0.62  & 0.93 & \textbf{0.84} & 0.87 & \textbf{0.86} & \textbf{0.94}  \\
        Llama-3.1 & 0.65 & \textbf{0.66} & \textbf{0.67} & \textbf{0.66} & \textbf{0.67} & \textcolor{blue}{\textbf{0.96}} & \textbf{0.84} & 0.85 & 0.84 & 0.85  \\ \hline

    \end{tabular}
    }

    \vspace{3mm}

    \setkeys{Gin}{keepaspectratio}
    \resizebox*{0.83\textwidth}{\textwidth} {
    \begin{tabular}{l|cccc|cccc}
    \Xhline{3\arrayrulewidth}
        \multicolumn{9}{c}{\textbf{\Large{Sarcasm}}}  \\
        \Xhline{3\arrayrulewidth}
        ~ & \multicolumn{4}{c|}{\textbf{Precision}} & \multicolumn{4}{c}{\textbf{Recall}}  \\  \hline
        Model & CM & CM+NHD & CM+iSarc & CM+SC-V2 & CM & CM+NHD & CM+iSarc & CM+SC-V2 \\ \hline \hline
        Gemma & 0.60 & \textbf{0.59} & 0.60 & 0.58  & 0.93 & 0.80 & 0.81 & 0.80 \\
        Aya Expanse &  0.66 & 0.52 & \textbf{0.66} & 0.66  & 0.93 & \textbf{0.94} & \textbf{0.95} & 0.93  \\
        Llama-3.1 & \textcolor{blue}{\textbf{0.67}} & 0.42 & 0.54 & \textbf{0.69} & \textcolor{blue}{\textbf{0.96}} & 0.75 & 0.93 & \textcolor{blue}{\textbf{0.96}}  \\  \hline

    \end{tabular}
    }

    \caption{Precision and recall scores of our experiment evaluating the impact of native language mixing in instruction fine-tuning of VMLMs using LoRA adapter. Notation: CM for code-mixed. The scores of best-performing models for the individual training scenarios are marked in bold and highest in the respective task is marked in \textcolor{blue}{blue}.}
    \label{tab:results_vmlm_lora_precall}
\end{table*}
\endgroup

\end{document}